\documentclass[times,twocolumn,final]{elsarticle}
\usepackage[table,dvipsnames]{xcolor}
\usepackage{medima}
\usepackage[utf8]{inputenc}
\usepackage[T1]{fontenc}
\usepackage{amsmath,amssymb,amsfonts,amsthm}
\usepackage{graphicx}
\usepackage{textcomp}
\usepackage{microtype}
\usepackage{booktabs,multirow,array}
\usepackage{subcaption}
\usepackage{placeins}
\usepackage{etoolbox}
\usepackage[hidelinks]{hyperref}
\biboptions{numbers,sort&compress}
\renewcommand{\titleboxawidth}{159.1mm}
\renewcommand{\titlegrayboxwidth}{131.8mm}
\renewcommand{\elsjnllogowidth}{20.4mm}
\renewcommand{\thefnote}{*}
\makeatletter
\patchcmd{\corref}{$\ast$}{$\dagger$}{}{\PackageError{authornotes}{Cannot set correspondence marker}{}}
\patchcmd{\cormark}{$\ast$}{$\dagger$}{}{\PackageError{authornotes}{Cannot set correspondence marker}{}}
\patchcmd{\cortext}{$\ast$}{$\dagger$}{}{\PackageError{authornotes}{Cannot set correspondence footnote marker}{}}
\makeatother

\newcommand{\best}[1]{\textbf{#1}}
\newcommand{\second}[1]{\underline{#1}}
\newcommand{\mstd}[2]{#1$_{\pm#2}$}
\newcommand{\sg}{\operatorname{sg}}
\newtheorem{proposition}{Proposition}
\newsavebox{\tablewidthbox}
\newcommand{\fitwidth}[2]{%
  \sbox{\tablewidthbox}{#2}%
  \ifdim\wd\tablewidthbox>#1\relax
    \resizebox{#1}{!}{\usebox{\tablewidthbox}}%
  \else\usebox{\tablewidthbox}\fi}

\journal{Medical Image Analysis}

\begin{document}
\verso{X. Li \textit{et~al.}}
\begin{frontmatter}
\title{Mitigating Spatial Blindness in Multiple Instance Learning for Whole-Slide Image Analysis}
\author[tju,shai]{Xiangyu Li\fnref{equal}}
\author[shai,miracle,bme]{Jincheng Li\fnref{equal}}
\author[shai,ntu]{Xing Xie}
\author[shai,ujn]{Bin Guan}
\author[bme,miracle]{Shaohua Kevin Zhou\corref{dagger}}
\author[bme,miracle,]{Rui Yan\corref{dagger}}
\author[tju]{Ran Su\corref{dagger}}
\address[tju]{College of Intelligence and Computing, Tianjin University, Tianjin, China}
\address[shai]{Shanghai Artificial Intelligence Laboratory, Shanghai, China}
\address[bme]{School of Biomedical Engineering, University of Science and Technology of China, Hefei, China}
\address[miracle]{Medical Imaging, Robotics, Analytic Computing \& Learning (MIRACLE) Lab, USTC Suzhou Institute for Advanced Research, Suzhou, China}
\address[ntu]{Engineering Training Center, Nantong University, Nantong, China}
\address[ujn]{School of Information Science and Engineering, University of Jinan, Jinan, China}
\fntext[equal]{Xiangyu Li and Jincheng Li contributed equally to this work and share first authorship.}
\cortext[dagger]{Corresponding authors: Shaohua Kevin Zhou, Rui Yan and Ran Su.}
\begin{abstract}
Whole-slide multiple instance learning can predict diagnostic and prognostic labels from tissue composition while making limited use of spatial organization. Providing coordinates or a graph does not establish that a trained predictor uses tissue neighborhoods. We examine this gap as spatial blindness and hypothesize that early composition learning can weaken supervision of a spatial pathway. We introduce ResTopoMIL, which freezes a learned composition predictor before training a graph residual with a coordinate-reassignment separation loss. Permutation and information analyses distinguish architectural symmetry, available spatial information and the proposed optimization mechanism. On a controlled spatial task, staged training achieved an AUC of 0.987, compared with 0.684 for the same jointly trained architecture. Performance remained high on a composition-only task. Across four WSI classification and five survival benchmarks, ResTopoMIL improved mean primary performance on eight tasks; four improvements remained significant after Holm-adjusted patient-cluster tests. Coordinate-free residual controls recovered some of the gains, while neighborhood destruction reduced performance close to the composition anchor. Neighborhood-preserving transformations and anatomy-stratified perturbations further characterized this dependence. These results support fixing a composition reference as a practical intervention for learning spatial corrections, while retaining additional appearance capacity as a contributing explanation.
\end{abstract}
\begin{keyword}
\KWD Computational pathology \sep Whole-slide images \sep Multiple instance learning \sep Spatial modeling
\end{keyword}
\end{frontmatter}
\section{Introduction}
Tissue composition and spatial organization provide complementary evidence for whole-slide prediction. Multiple instance learning (MIL) addresses this task using bags of patch features and slide-level labels \citep{dietterich1997solving,maron1997framework,campanella2019clinical,lu2021data}. This weak supervision makes large histopathology images tractable without dense annotations \citep{pantanowitz2011review,verghese2023computational,song2023artificial}. However, a high diagnostic or prognostic score does not reveal which tissue evidence supports the prediction. Tumor abundance can be informative, whereas gland formation, invasive fronts and tumor--stroma interfaces depend on spatial relationships \citep{quail2013microenvironmental,cheng2021computational,wang2021predicting}.

\begin{figure*}[!t]
    \centering
    \includegraphics[width=\textwidth]{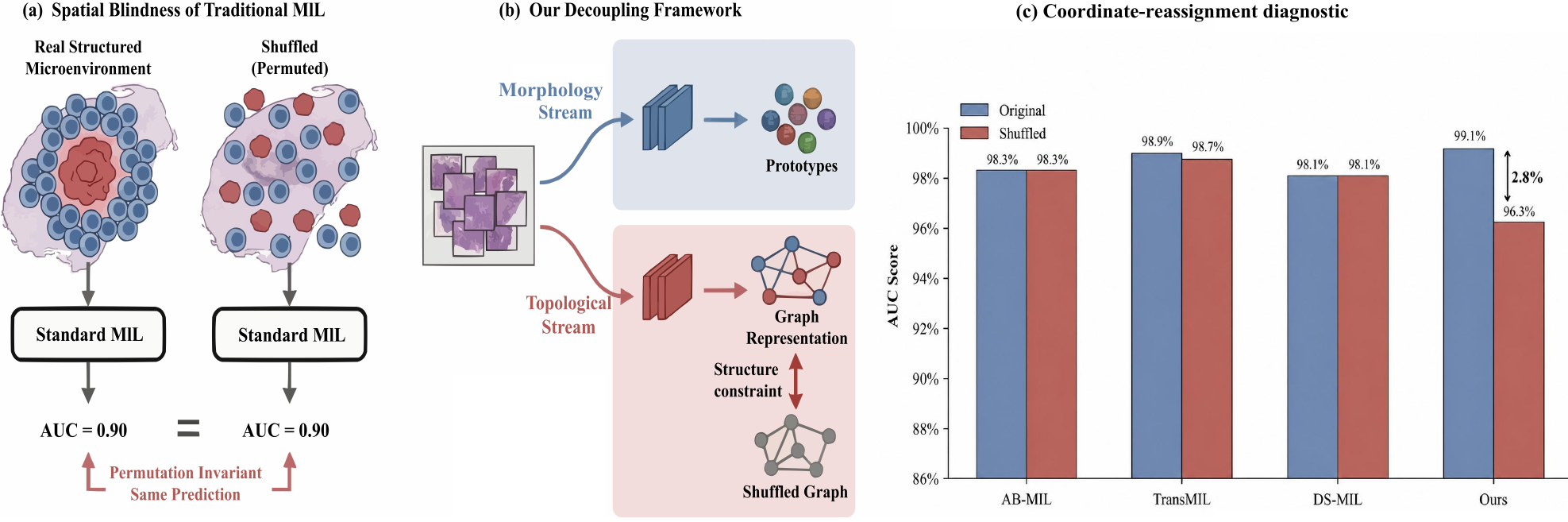}
    \caption{Motivation and coordinate-reassignment diagnostic. (a) Coordinate-free MIL receives the same appearance multiset after coordinate reassignment and therefore gives the same prediction; joint relabeling of paired features and coordinates is a different operation. The schematic AUC values are illustrative. (b) ResTopoMIL separates a composition reference from a graph correction constrained to distinguish intact from disrupted neighborhoods. (c) The TCGA-BRCA diagnostic holds patch embeddings fixed while reassigning coordinates. Near-invariance is expected when a model's effective inputs do not change; it does not itself establish failed spatial learning. Table~\ref{tab:shuffle_endpoint_main} tests spatial baselines with rebuilt coordinate-derived inputs.}
    \label{fig:teaser}
\end{figure*}

Accurate prediction from an unordered bag therefore leaves an unresolved question about spatial learning. Attention pooling is invariant to instance order, and self-attention without positional encoding is equivariant before invariant readout \citep{ilse2018attention,zaheer2017deep,chen2025jigsaw}. These properties allow rich appearance statistics to support prediction without physical coordinates. They do not imply that every order-invariant model ignores space. A graph predictor can remain invariant to node relabeling while depending on its adjacency. The relevant intervention changes feature--coordinate associations, rather than merely changing the order in which paired instances are listed.

Spatial architectures provide access to neighborhoods, but access alone does not establish learned use. Graph networks, positional Transformers, hierarchical aggregators and state-space models introduce spatial or contextual pathways \citep{chen2021whole,pati2022hierarchical,adnan2020representation,shao2021transmil,chen2022scaling,yang2024mambamil,zhang20252dmamba}. Their predictors can still exploit appearance signals that are easier to learn. We use \textit{spatial blindness} to describe weak prediction or performance response to neighborhood destruction on a structure-dependent task. Figure~\ref{fig:teaser}(c) motivates this behavioral distinction, while comparisons with spatial baselines explicitly rebuild their coordinate-derived inputs. The unresolved gap concerns how training can increase useful neighborhood dependence beyond simply adding another prediction branch.

Our optimization hypothesis is that composition can reduce prediction error before a spatial pathway learns useful relations. For additive logits, the supervised graph update is multiplied by the model's current prediction error. This creates a conditional route to weak spatial supervision, consistent with shortcut-learning and gradient-starvation accounts \citep{shah2020pitfalls,geirhos2018imagenet,noroozi2016unsupervised,pezeshki2021gradient}. The gradient identity motivates an intervention; it does not prove that composition always learns first or that freezing must increase gradients.

ResTopoMIL fixes a composition reference before learning a graph-conditioned correction (Figure~\ref{fig:teaser}). A prototype histogram first predicts from the appearance multiset. After this predictor is frozen, a graph branch learns residual logits, while a margin loss separates intact and coordinate-reassigned graph representations. This design makes the training schedule testable within the same architecture and makes neighborhood dependence testable independently of appearance capacity.

\begin{samepage}
The contributions of this study are threefold:
\begin{itemize}
    \item We formulate spatial blindness as weak neighborhood dependence in WSI MIL and distinguish it from permutation symmetry. A conditional gradient analysis motivates the proposed optimization hypothesis.
    \item We introduce ResTopoMIL, which freezes a prototype-based composition predictor before learning an additive graph residual. A coordinate-reassignment separation loss encourages the graph branch to distinguish intact from disrupted neighborhoods.
    \item We evaluate the intervention using controlled composition and spatial tasks, nine WSI prediction benchmarks and matched ablations. Coordinate-free residual controls, neighborhood perturbations and anatomy-stratified tests distinguish spatial dependence from additional appearance capacity.
\end{itemize}
\end{samepage}

\section{Related work}
\subsection{Appearance aggregation under slide-level supervision}
MIL learns bag predictions without requiring a label for every instance. Early pathology methods combined patch classifiers, clustering, global pooling or multi-view CNN representations \citep{xu2014weakly,kraus2016classifying,das2017classifying,das2018multiple,wang2019weakly,chen2021annotation}. Attention MIL learns instance weights within an order-invariant readout \citep{ilse2018attention,lu2021data}. Later approaches use dual streams, bag distillation, hard-instance mining and attention regularization to improve discrimination \citep{li2021dual,zhang2022dtfd,tang2023multiple,lin2023interventional,zhang2024attention,zhang2025aem,qu2024rethinking,shao2025multiple,zhu2025effective,zhu2023pdl}. These methods show that appearance aggregation is a strong predictive baseline, making its contribution essential to isolate when evaluating spatial models.

Stronger patch encoders increase the information available to any aggregator. Pathology foundation models and self-supervised representations improve local embeddings \citep{chen2023general,kang2023benchmarking,oquab2023dinov2,caron2021emerging,lu2023visual,kapse2025gecko}. Vision-language and prompt-based methods also broaden slide supervision \citep{shi2024vila,han2025mscpt,wongfew,tomar2025slide,gou2025queryable}. We therefore hold the patch encoder fixed within each comparison and evaluate classification with both UNI and CTransPath. This protocol targets aggregation and training rather than additional encoder capacity.

\subsection{Spatial and contextual aggregation}
Structured MIL recognizes that relations between instances can affect bag labels \citep{zhou2009multi,zhang2011multiple}. Graph methods represent tissue as a point cloud or neighborhood graph \citep{chen2021whole,adnan2020representation,pati2022hierarchical,pal2022bag}. Transformers and low-rank attention model interactions across large bags \citep{vaswani2017attention,shao2021transmil,xiong2021nystromformer,chen2022scaling,xiang2023exploring}. Hierarchical and context-aware methods exploit WSI pyramids and regional organization \citep{zhang2021joint,guo2023higt,buzzard2024paths,tran2025navigating,fourkioticamil,chen2024camil}, while state-space models address long-sequence computation \citep{guefficiently,yang2024mambamil,zhang20252dmamba}. Our question complements these architectural developments by testing whether a trained predictor uses the spatial input it receives.

\subsection{Spatial learning and optimization}
Prior work has explicitly examined order symmetry and positional information. Chen et al.~\citep{chen2025jigsaw} analyzed these properties and proposed restoring instance arrangement through a Siamese formulation. NcIEMIL addressed feature-extraction and aggregation redundancy using filtering and spatial/channel attention \citep{sun2024nciemil}. ResTopoMIL instead tests a training intervention when appearance and graph pathways coexist. Its permutation arguments build on established set-model analysis, rather than constituting a new general theory of MIL.

The intervention combines established representation ideas with a different optimization schedule. Deep Sets, histograms and morphological prototypes motivate the composition reference \citep{zaheer2017deep,peeples2021histogram,song2024morphological,wei2016scalable}. Shortcut learning motivates testing whether easy signals dominate joint training \citep{shah2020pitfalls,pezeshki2021gradient}. Conditional-information identities describe what remains beyond a fixed reference \citep{shannon1948mathematical,cover1999elements}. The contribution lies in connecting these ideas through a staged model and evaluating its neighborhood dependence under matched controls.
\section{Problem formulation and analysis}
\label{sec:prelim_motivation}
The analysis separates three questions: what a model can represent, what spatial information is available, and what optimization may encourage it to learn. This distinction prevents permutation symmetry from being treated as proof of spatial blindness.

\subsection{Permutation symmetry and neighborhood dependence}
A slide is a paired bag $X=\{(\mathbf h_i,\mathbf p_i)\}_{i=1}^{N}$, where $\mathbf h_i\in\mathbb R^d$ is a patch feature and $\mathbf p_i\in\mathbb R^2$ its coordinate. Let $\mathbf H$ and $\mathbf R$ stack these quantities by row. The appearance multiset is $B=\{\mathbf h_i\}_{i=1}^{N}$. Composition denotes statistics of $B$, whereas spatial organization depends on the feature--coordinate pairing.

\begin{proposition}[Permutation symmetry does not imply spatial blindness]
\label{prop:permutation}
Let $\Pi$ be a permutation matrix and $\rho$ an invariant readout. An equivariant coordinate-free map $T$ satisfies $\rho(T(\Pi\mathbf H))=\rho(T(\mathbf H))$. An equivariant graph map $G$ satisfies
\begin{equation}
\rho\big(G(\Pi\mathbf H,\Pi\mathbf A\Pi^\top)\big)
=\rho\big(G(\mathbf H,\mathbf A)\big).
\label{eq:joint_relabeling}
\end{equation}
Neither identity requires invariance when $\mathbf A$ changes while $\mathbf H$ remains fixed.
\end{proposition}

\begin{proof}
Attention pooling provides a coordinate-free example. For a shared scorer $a$, define $e_i=a(\mathbf h_i)$ and $\boldsymbol\alpha(\mathbf H)=\mathrm{softmax}([e_1,\ldots,e_N]^\top)$. Its pooled representation obeys
\begin{equation}
\begin{aligned}
z(\Pi\mathbf H)
&=(\Pi\mathbf H)^\top\Pi\boldsymbol\alpha(\mathbf H)\\
&=\mathbf H^\top\boldsymbol\alpha(\mathbf H)=z(\mathbf H),
\end{aligned}
\end{equation}
because $\Pi^\top\Pi=\mathbf I$. Shared gated scorers have the same property \citep{ilse2018attention}.

Self-attention without positional encoding is equivariant before pooling. Let $\mathbf Q=\mathbf H\mathbf W_Q$, $\mathbf K=\mathbf H\mathbf W_K$ and $\mathbf V=\mathbf H\mathbf W_V$. Row-wise softmax satisfies $\mathrm{softmax}(\Pi\mathbf M\Pi^\top)=\Pi\mathrm{softmax}(\mathbf M)\Pi^\top$, including permutation of each row's normalization sum. Consequently,
\begin{equation}
\begin{aligned}
T(\Pi\mathbf H)
&=\mathrm{softmax}\!\left(\frac{\Pi\mathbf Q\mathbf K^\top\Pi^\top}{\sqrt{d_k}}\right)\Pi\mathbf V\\
&=\Pi T(\mathbf H).
\end{aligned}
\end{equation}
Shared tokenwise feed-forward maps, activations and layer normalization preserve equivariance. Invariant pooling gives an invariant bag predictor. A class token is handled with $\bar\Pi=\mathrm{diag}(1,\Pi)$, which leaves its output row unchanged \citep{chen2025jigsaw}.

For a GCN, let $\hat{\mathbf A}=\tilde{\mathbf D}^{-1/2}(\mathbf A+\mathbf I)\tilde{\mathbf D}^{-1/2}$. Joint relabeling gives $\mathbf H'=\Pi\mathbf H$ and $\hat{\mathbf A}'=\Pi\hat{\mathbf A}\Pi^\top$. Each shared-weight layer satisfies
\begin{equation}
\sigma(\hat{\mathbf A}'\mathbf H'\mathbf W)
=\Pi\sigma(\hat{\mathbf A}\mathbf H\mathbf W).
\end{equation}
Induction establishes stack equivariance, and invariant readout gives Eq.~\eqref{eq:joint_relabeling}.
\end{proof}

Coordinate reassignment tests a different property from joint relabeling. For an equivariant graph constructor with consistently resolved distance ties, $\mathbf A(\Pi\mathbf R)=\Pi\mathbf A(\mathbf R)\Pi^\top$. Reindexing $(\mathbf H,\mathbf R)$ together preserves the paired bag. Holding $\mathbf H$ fixed and permuting only $\mathbf R$ instead changes which appearances occupy adjacent locations. The coordinate set and unlabeled geometric graph remain unchanged, but feature--neighborhood associations differ. Proposition~\ref{prop:permutation} imposes no equality between the resulting predictions. These symmetry statements assume deterministic evaluation and do not describe stochastic dropout differences.

\subsection{Positional access does not ensure learned sensitivity}
Positional encoding can remove an architectural invariance guarantee without guaranteeing different predictions. For an invariant aggregator $F$ and slot-attached encodings $\mathbf E$,
\begin{equation}
F(\Pi\mathbf H+\mathbf E)=F(\mathbf H+\Pi^\top\mathbf E).
\end{equation}
The right-hand side need not equal $F(\mathbf H+\mathbf E)$. However, distinct inputs can share an output. Mean pooling gives the counterexample
\begin{equation}
\frac1N\mathbf1^\top(\Pi\mathbf H+\mathbf E)
=\frac1N\mathbf1^\top(\mathbf H+\mathbf E).
\end{equation}
Repeated encoding rows can also admit nontrivial invariant permutations. Conversely, jointly permuting features and their encodings preserves the output of any invariant $F$.

Sensitivity must therefore be evaluated for the trained model and the intervention actually supplied. Physical-coordinate reassignment may leave a padded positional grid unchanged. Such a model is an input-preserving control in that test, rather than evidence that its positional module failed to learn. Spatial baselines are evaluated by rebuilding their graph or coordinate-derived sequence.

\subsection{Spatial information beyond appearance}
An appearance predictor can be accurate even when coordinates contain additional label information. Since $B$ is a deterministic projection of $X$, standard conditional-information identities yield
\begin{equation}
\begin{aligned}
I(Y;X\mid B)
&=H(Y\mid B)-H(Y\mid X,B)\\
&=H(Y\mid B)-H(Y\mid X)\geq0.
\end{aligned}
\label{eq:spatial_information}
\end{equation}
For discrete labels, this also equals $\mathbb E_X\mathrm{KL}(p(Y\mid X)\|p(Y\mid B))$ \citep{chen2025jigsaw,cover1999elements}. It vanishes precisely when these posteriors agree almost surely.

Positive conditional information establishes a potential reduction in optimal expected log loss. It does not require a strict reduction in zero--one classification error, because the most probable class may remain unchanged. A predictor with access to $X$ can always ignore coordinates, so its optimal classification risk cannot exceed that using $B$. Neither statement guarantees that a finite-data network learns the available spatial signal. We reserve spatial blindness for weak behavioral dependence on neighborhoods on a task where those neighborhoods matter.

\subsection{A conditional optimization hypothesis}
\label{sec:theoretical_insights}
An additive predictor exposes one route by which composition may weaken spatial supervision. Write
\begin{equation}
f(X)=f_{stat}(B;\theta_s)+f_{topo}(X;\theta_t).
\label{eq:additive_decomp}
\end{equation}
This logit decomposition does not identify separate appearance and topology factors. The graph term can also encode appearance information omitted by the statistical predictor.

\begin{proposition}[Residual-error gating of the graph update]
\label{prop:residual_gradient}
For binary cross-entropy, let $r_\theta(X,Y)=\sigma(f(X))-Y$. Assume finite second moments and valid interchange of differentiation and expectation. Then
\begin{equation}
\nabla_{\theta_t}\mathcal L
=\mathbb E\big[r_\theta(X,Y)\nabla_{\theta_t}f_{topo}(X)\big],
\label{eq:main_topo_gradient}
\end{equation}
and
\begin{equation}
\|\nabla_{\theta_t}\mathcal L\|
\leq\sqrt{\mathbb E r_\theta^2}\,
\sqrt{\mathbb E\|\nabla_{\theta_t}f_{topo}\|_F^2}.
\label{eq:main_gradient_bound}
\end{equation}
\end{proposition}
\begin{proof}
Set $\hat p_\theta=\sigma(f(X))$. Differentiating the binary loss $-Y\log\hat p_\theta-(1-Y)\log(1-\hat p_\theta)$ with respect to $f$ gives $\hat p_\theta-Y$. The chain rule gives Eq.~\eqref{eq:main_topo_gradient}. The norm inequality and Cauchy--Schwarz give Eq.~\eqref{eq:main_gradient_bound}.
\end{proof}

The bound contracts if composition reduces the squared prediction error early while the graph Jacobian remains bounded. This condition motivates testing whether a fixed composition reference improves subsequent graph learning. It does not establish that composition always learns first or compare staged and joint gradient magnitudes. The toy experiment tests the schedule within one architecture, while gradient traces and WSI controls examine its empirical consequences.
\section{ResTopoMIL}
\label{sec:method}
\begin{figure*}[!t]
\centering
\includegraphics[width=\textwidth]{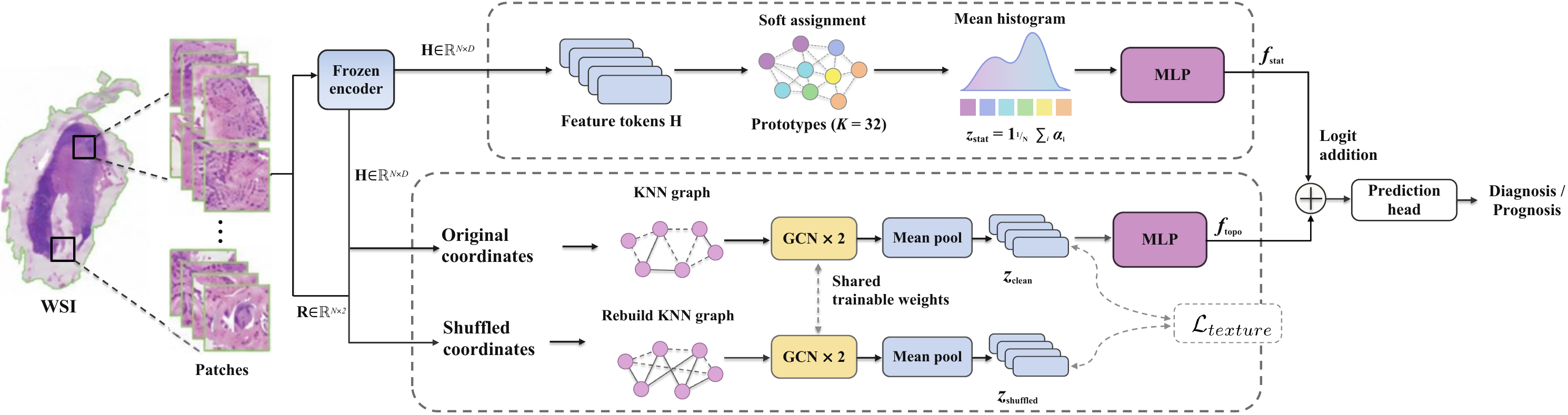}
\caption{ResTopoMIL framework. A frozen patch encoder supplies shared appearance features. The statistical stream averages soft prototype assignments into a composition histogram and is trained in Stage 1, then frozen in Stage 2. The topological stream constructs KNN graphs from intact and reassigned coordinates and processes both views with a shared, trainable two-layer GCN. A margin loss separates their pooled representations. The statistical and intact-graph prediction logits are added before probability conversion, as specified in Eq.~\eqref{eq:fusion}. The reassigned view and separation loss are used only during training; inference uses the intact graph.}
\label{fig:arch}
\end{figure*}

ResTopoMIL separates a fixed composition reference from a trainable graph correction. Figure~\ref{fig:arch} shows the three design choices needed for this intervention: an order-invariant anchor, staged additive prediction and a coordinate-specific representation constraint. The constraint tests graph dependence beyond the residual objective itself.

\subsection{A prototype histogram as the composition reference}
\label{sec:stat_stream}
The statistical stream estimates how much the appearance multiset explains without coordinates. A constrained prototype histogram provides an inspectable reference without assuming that it exhausts all appearance information. Let $\mathcal C=\{\mathbf c_k\}_{k=1}^{K_{proto}}$ be a learnable codebook. For each patch, the soft assignment is
\begin{equation}
a_{ik}=\frac{\exp(-\|\mathbf h_i-\mathbf c_k\|^2/\tau)}
{\sum_{j=1}^{K_{proto}}\exp(-\|\mathbf h_i-\mathbf c_j\|^2/\tau)}.
\label{eq:assignment}
\end{equation}
The temperature $\tau$ controls assignment sharpness. Averaging gives the bag histogram and statistical logits,
\begin{equation}
\mathbf z_{stat}=\frac1N\sum_i\mathbf a_i,
\qquad f_{stat}=\mathrm{MLP}(\mathbf z_{stat}).
\end{equation}
The histogram records soft prototype proportions, rather than diagnostic class probabilities or spatial locations. Its entries sum to one because each patch assignment is normalized.

Prototype frequencies summarize recurring appearance patterns \citep{wei2016scalable}. MiniBatch K-Means initializes the codebook from training patch embeddings, after which both the prototypes and temperature are learned. A 512-dimensional hidden representation precedes the statistical prediction head. Anchor-convergence tests examine the warm-up duration, and coordinate-free residual controls quantify usable appearance information omitted by this reference.

\subsection{A graph-conditioned correction}
\label{sec:topo_stream}
The graph stream gives the correction access to physical neighborhoods while retaining the same patch features. Each patch forms a node, and a coordinate KNN graph supplies the adjacency $\mathbf A$. We use two GCN layers,
\begin{equation}
\mathbf H^{(l+1)}=\sigma\!\left(
\tilde{\mathbf D}^{-1/2}\tilde{\mathbf A}
\tilde{\mathbf D}^{-1/2}\mathbf H^{(l)}\mathbf W^{(l)}\right),
\end{equation}
where $\tilde{\mathbf A}=\mathbf A+\mathbf I$, $\tilde{\mathbf D}$ is its degree matrix and $\mathbf H^{(0)}=\mathbf H$. Global mean pooling and a linear head give
\begin{equation}
\mathbf z_{topo}=\frac1N\sum_i\mathbf H_i^{(2)},
\qquad f_{topo}=\mathbf W_{topo}\mathbf z_{topo}+\mathbf b_{topo}.
\end{equation}

Coordinate neighborhoods distinguish visually similar patches occupying different tissue contexts. The default graph has eight nearest neighbors per node before self-loops, and its hidden features have dimension 512. ReLU and dropout with probability 0.25 follow the graph transformations. The shallow branch remains fixed in the joint-versus-staged comparison, allowing the training schedule to be evaluated without changing architecture.

\subsection{Additive logits with a fixed reference}
\label{sec:optimization}
Fixing the anchor prevents the composition predictor from changing throughout residual learning. Stage 1 trains the statistical stream alone for ten epochs. Stage 2 freezes its codebook, temperature, MLP and prediction head, then optimizes the graph branch through
\begin{equation}
f(X)=\sg[f_{stat}(B)]+f_{topo}(X),
\label{eq:fusion}
\end{equation}
where $\sg$ stops gradients. Both heads produce logits in the same output space. Classification probabilities are obtained only after summation, using Softmax for multiclass prediction or a sigmoid for a single binary logit.

The residual is defined by the predictive objective, rather than by subtracting hidden features or fitting a separate residual label. The supervised error evolves as the graph learns, but the anchor cannot absorb changes in that error. Larger graph learning rates, statistical dropout, curricula and hard-instance mining alter joint optimization while leaving the reference trainable. Table~\ref{tab:ablation_main} compares these interventions. Freezing changes the learning problem, although Proposition~\ref{prop:residual_gradient} provides no guarantee that its gradients are larger.

\subsection{Separating intact and reassigned neighborhoods}
Residual prediction alone can recover appearance information that the anchor missed. We therefore use a coordinate-specific constraint to discourage the graph stream from ignoring its adjacency. A corrupted bag $\tilde X$ retains $\mathbf H$ and reassigns coordinates as $\Pi\mathbf R$, after which the KNN graph is rebuilt. The same trainable GCN processes both views and produces $\mathbf z_{topo}$ and $\tilde{\mathbf z}_{topo}$.

The auxiliary objective separates these representations using a cosine-distance margin,
\begin{equation}
\mathcal L_{texture}
=\max\!\left(0,\,m-\left[1-\mathrm{cos}(\mathbf z_{topo},\tilde{\mathbf z}_{topo})\right]\right).
\label{eq:texture}
\end{equation}
The loss decreases as similarity falls, and vanishes once the cosine distance reaches $m$. At the default $m=0.3$, the zero-loss condition is similarity at most 0.7. This is a separation constraint without an explicit positive-pair attraction term. Coordinate reassignment is not assumed to preserve the tissue label.

Supervision uses the intact view, while the reassigned view enters only the representation comparison. The Stage-2 classification objective is
\begin{equation}
\mathcal L_{total}
=\mathcal L_{cls}(\sg[f_{stat}]+f_{topo},Y)
+\lambda\mathcal L_{texture}.
\label{eq:total_loss}
\end{equation}
Survival prediction replaces classification loss with the discrete-time survival objective. A deterministic graph-ignoring representation cannot distinguish the two views, but representation separation alone does not ensure that prediction uses the difference. Inference therefore uses only the intact graph and sums the two logits once; perturbation experiments test the learned predictor's dependence on neighborhoods.

\subsection{Information captured beyond the anchor}
The residual objective targets information beyond a fixed reference without identifying that information as spatial. Let $Z_{stat}$ and $Z_{topo}$ denote the branch representations as random variables. The chain rule gives
\begin{equation}
I(Z_{stat},Z_{topo};Y)
=I(Z_{stat};Y)+I(Z_{topo};Y\mid Z_{stat}).
\label{eq:mi_chain}
\end{equation}
Stage 2 fixes the first term, while the graph branch and its head parameterize
\begin{equation}
q_\phi(y\mid Z_{stat},Z_{topo})
=\left[\mathrm{Softmax}(f_{stat}+f_{topo,\phi})\right]_y.
\label{eq:decoder}
\end{equation}
For binary prediction, the positive-class probability is $\sigma(f_{stat}+f_{topo})$.

\begin{proposition}[Residual conditional-information bound]
\label{prop:residual_information}
With $Z_{stat}$ and $f_{stat}$ fixed, minimizing the decoder's cross-entropy maximizes the variational lower bound
\begin{equation}
\begin{aligned}
I(Z_{topo};Y\mid Z_{stat})
&\geq H(Y\mid Z_{stat})\\
&\quad+\mathbb E\log q_\phi(Y\mid Z_{stat},Z_{topo}).
\end{aligned}
\label{eq:main_mi_chain}
\end{equation}
\end{proposition}
\begin{proof}
Let $Z=(Z_{stat},Z_{topo})$. The decoder's expected cross-entropy decomposes as
\begin{equation}
\begin{aligned}
&\mathbb E[-\log q_\phi(Y\mid Z)]\\
&\quad=H(Y\mid Z)+\mathbb E_Z\mathrm{KL}(p(Y\mid Z)\|q_\phi(Y\mid Z)).
\end{aligned}
\end{equation}
Nonnegative KL divergence bounds the conditional entropy by cross-entropy. Substitution into $I(Z_{topo};Y\mid Z_{stat})=H(Y\mid Z_{stat})-H(Y\mid Z)$ gives Eq.~\eqref{eq:main_mi_chain}. The first term is constant during Stage 2, so maximizing this lower bound is equivalent to minimizing the expected negative log likelihood.
\end{proof}

Increasing the bound does not itself guarantee an increase in true conditional mutual information. The graph representation can also encode omitted appearance statistics. Equation~\eqref{eq:texture} supplies a separate spatial constraint, while coordinate-free residuals, graph perturbations and anatomical tests assess the interpretation empirically.
\section{Experiments}
\label{sec:experiments}
The experiments test the argument at three levels. Controlled bags isolate composition from spatial labels, WSI benchmarks measure predictive utility, and matched ablations and perturbations examine the source of the gains. This order separates successful prediction from evidence of neighborhood use.

\subsection{Evaluation protocol}
\subsubsection{WSI datasets and prediction tasks}
Nine public WSI benchmarks cover diagnostic classification and survival prediction. TCGA-NSCLC contains 1,053 slides, comprising 541 LUAD and 512 LUSC cases, for lung cancer subtyping. TCGA-BRCA contains 1,021 slides for invasive ductal versus invasive lobular carcinoma, contrasting ductal architecture with single-file infiltration patterns. BRACS uses 525 slides in three classes: benign, atypical and malignant \citep{brancati2022bracs}. PANDA contains 10,616 slides for Gleason grading, where gland formation and fused-gland architecture are relevant tissue features.

Five TCGA cohorts test survival prediction: KIRC (519 slides), KIRP (259), LUAD (516), STAD (441) and UCEC (537). The cohorts span diverse morphological factors without assuming that all prognostic information is spatial. LUAD includes lepidic, acinar, papillary, micropapillary and solid growth patterns. STAD includes gland differentiation, structural abnormalities and signet-ring morphology, while UCEC includes variation in the glandular-to-solid component ratio. KIRC includes nuclear-grade variation, and KIRP includes papillary-core organization. Models predict discrete time-to-event risk over four non-overlapping intervals. CAMELYON-16 provides a separate localization evaluation using tumor masks and is not counted among the nine prediction benchmarks.

\subsubsection{Controlled composition and spatial tasks}
Spatial-MNIST-Bag isolates the training hypothesis without pathology-specific annotation. Each bag contains 50 digit instances represented by 512-dimensional pretrained ResNet-18 embeddings and assigned coordinates in $[0,1]^2$. Dataset A samples coordinates independently from $U([0,1]^2)$ and labels a bag positive if it contains at least one digit ``9''. Coordinates therefore provide no systematic label information.

Dataset B matches the digit-composition distribution across classes and varies their spatial arrangement. Each bag contains one each of ``1,3,5,7,9'' and 45 even digits sampled from $\{0,2,4,6,8\}$. Background digits and individual embeddings vary across bags according to the same class-independent sampling rule. Positive bags draw a centroid $c\sim U([0.2,0.8]^2)$ and place the five odd digits around it using Gaussian noise with $\sigma=0.05$. Background coordinates remain uniform. Negative bags scatter all 50 instances uniformly. The positive construction encourages a compact odd-digit neighborhood; Gaussian sampling does not guarantee a particular KNN clique.

\subsubsection{Preprocessing, baselines and implementation}
All comparisons within an encoder setting share patch features, patient-level splits and evaluation procedures. Patient-level stratification prevents slides from the same patient appearing in training, validation and test partitions. Tissue is identified using Otsu thresholding in HSV space and divided into non-overlapping $256\times256$ patches at $20\times$ magnification. The primary evaluation uses frozen 1024-dimensional UNI embeddings. UNI uses a ViT-Large architecture pretrained with DINOv2 on Mass-100K, comprising more than $10^8$ patches. A separate classification comparison uses frozen CTransPath features to assess consistency across encoders.

The classification and survival tables compare ResTopoMIL with AB-MIL, CLAM-SB, DS-MIL, TransMIL, ILRA-MIL, MHIM-MIL, DGR-MIL and 2DMambaMIL. DTFD-MIL is additionally included in the localization comparison. PatchGCN supplies a graph baseline in the coordinate-reassignment diagnostic. Optimizer family, preprocessing and evaluation protocols are matched within the reported comparisons.

The default model uses $K_{proto}=32$, $K_{knn}=8$, $m=0.3$ and $\lambda=1.0$. Prototype initialization uses MiniBatch K-Means on a random subset of training patches; the learnable temperature starts at 1.0. After ten statistical warm-up epochs, refinement trains the graph branch for 30 classification epochs or 20 survival epochs. Optimization uses Lookahead with an Adam base, learning rate $2\times10^{-4}$, weight decay $10^{-4}$ and cosine annealing. Experiments use PyTorch 1.13, PyTorch Geometric and one NVIDIA RTX 3090 with 24 GB memory.

\subsubsection{Metrics and statistical analysis}
Classification is assessed with accuracy, F1 and AUC; survival prediction uses the concordance index. Unless a caption states otherwise, tables report means and standard deviations over five matched runs. Representative trajectories and endpoints without run-level dispersion are identified separately. Bold and underlined values rank reported means rather than statistical significance.

Patient-cluster inference compares each primary UNI endpoint with the baseline having the highest original mean AUC or C-index. All slides and repeated appearances of a patient are grouped, with shared weights and method-label swaps across methods and runs. We use 10,000 bootstrap samples for marginal 95\% confidence intervals and 100,000 permutations for two-sided $p=(b+1)/(B+1)$. Holm adjustment covers the nine primary endpoints. These intervals quantify patient-sampling uncertainty, while seed standard deviations describe run-to-run variability.

\subsection{Staged training recovers a controlled spatial rule}
\label{sec:synthetic_main}
The toy tasks test whether the training schedule improves spatial learning while retaining appearance-based prediction. Dataset A rewards instance detection, whereas Dataset B removes a systematic digit-composition rule. Table~\ref{tab:synthetic_double_dissociation} reports the resulting AUCs.

\begin{table}[!t]

    \centering
    \footnotesize
    \caption{Controlled composition and spatial tasks. Dataset A requires instance detection without useful coordinates; Dataset B matches digit composition and varies spatial arrangement. Bold and underlined AUCs denote the highest and second-highest reported values per task. Run-level dispersion is unavailable for these endpoints.}
    \label{tab:synthetic_double_dissociation}

    \setlength{\tabcolsep}{3.2pt}
    \renewcommand{\arraystretch}{1.05}
    \fitwidth{\linewidth}{
    \begin{tabular}{l cc}
        \toprule
        \textbf{Method} & \textbf{A: Comp.} & \textbf{B: Topo.} \\
        & \textbf{AUC} & \textbf{AUC} \\
        \midrule
        AB-MIL & \best{0.998} & 0.505 \\
        TransMIL & \second{0.995} & 0.532 \\
        ResTopoMIL (Joint) & 0.991 & \second{0.684} \\
        \midrule
        \textbf{ResTopoMIL} & 0.994 & \best{0.987} \\
        \bottomrule
    \end{tabular}}

\end{table}

On Dataset B, staged ResTopoMIL achieved AUC 0.987, compared with 0.684 for the same architecture trained jointly. AB-MIL and TransMIL were near chance at 0.505 and 0.532. Both performed well on Dataset A, with AUCs of 0.998 and 0.995, respectively. ResTopoMIL retained AUC 0.994 on that composition task, although it did not exceed AB-MIL. The matched-architecture contrast isolates the schedule's contribution to learning the controlled spatial label.

\subsection{Predictive utility across WSI classification tasks}
The four classification benchmarks test whether the intervention has value beyond synthetic bags. Table~\ref{tab:classification_main} compares aggregation methods under two frozen encoders, holding the feature source fixed within each block.

\begin{table*}[!t]
    \centering
    \scriptsize
    \caption{WSI classification with frozen UNI and CTransPath (CPath) features. Entries are mean$\pm$standard deviation over five runs. Params is the trainable parameter count in millions. Bold and underlined means mark the best and second-best distinct values within each encoder block; smaller parameter counts are preferred. Ties share formatting. Rankings do not establish significance.}
    \label{tab:classification_main}
    \setlength{\tabcolsep}{1.25pt}
    \renewcommand{\arraystretch}{1.01}
    \fitwidth{\textwidth}{
    \begin{tabular}{llc ccc ccc ccc ccc}
        \toprule
        \multirow{2}{*}{\textbf{Feature}} & \multirow{2}{*}{\textbf{Method}} & \multirow{2}{*}{\textbf{Params}} &
        \multicolumn{3}{c}{\textbf{BRACS}} &
        \multicolumn{3}{c}{\textbf{PANDA}} &
        \multicolumn{3}{c}{\textbf{TCGA-NSCLC}} &
        \multicolumn{3}{c}{\textbf{TCGA-BRCA}} \\
        \cmidrule(lr){4-6} \cmidrule(lr){7-9} \cmidrule(lr){10-12} \cmidrule(lr){13-15}
        & & & Acc & F1 & AUC & Acc & F1 & AUC & Acc & F1 & AUC & Acc & F1 & AUC \\
        \midrule
        \multirow{9}{*}{\textbf{UNI}} & AB-MIL & \best{.59} & \mstd{.7275}{.0759} & \mstd{.6748}{.0185} & \mstd{.8806}{.0091} & \mstd{.7322}{.0059} & \mstd{.6876}{.0079} & \mstd{.9306}{.0017} & \mstd{.8988}{.0066} & \mstd{.8680}{.0068} & \mstd{.9569}{.0073} & \mstd{.9414}{.0062} & \mstd{.9187}{.0106} & \mstd{.9727}{.0019} \\
        & CLAM-SB & \second{.79} & \mstd{\second{.7371}}{.0182} & \mstd{\second{.7051}}{.0287} & \mstd{\second{.8840}}{.0131} & \mstd{.7318}{.0041} & \mstd{.6814}{.0050} & \mstd{.9215}{.0009} & \mstd{.8836}{.0085} & \mstd{.8529}{.0084} & \mstd{.9635}{.0064} & \mstd{.9314}{.0062} & \mstd{\second{.9268}}{.0112} & \mstd{\second{.9814}}{.0027} \\
        & DS-MIL & 1.20 & \mstd{.6460}{.0189} & \mstd{.6068}{.0436} & \mstd{.8054}{.0225} & \mstd{.7394}{.0207} & \mstd{\second{.6911}}{.0231} & \mstd{\second{.9309}}{.0070} & \mstd{.8836}{.0085} & \mstd{.8627}{.0099} & \mstd{.9579}{.0083} & \mstd{.9409}{.0124} & \mstd{.9058}{.0208} & \mstd{.9777}{.0039} \\
        & TransMIL & 2.67 & \mstd{.6506}{.0174} & \mstd{.5592}{.0398} & \mstd{.8450}{.0098} & \mstd{.7090}{.0088} & \mstd{.6667}{.0133} & \mstd{.9288}{.0038} & \mstd{.8933}{.0132} & \mstd{.8822}{.0132} & \mstd{.9692}{.0084} & \mstd{.9409}{.0095} & \mstd{.9016}{.0163} & \mstd{.9787}{.0154} \\
        & ILRA-MIL & 3.68 & \mstd{.6230}{.0286} & \mstd{.5526}{.0368} & \mstd{.8012}{.0219} & \mstd{\second{.7402}}{.0096} & \mstd{.6880}{.0094} & \mstd{.9266}{.0034} & \mstd{.8912}{.0157} & \mstd{.8829}{.0159} & \mstd{.9628}{.0095} & \mstd{.9455}{.0051} & \mstd{.9108}{.0091} & \mstd{.9621}{.0148} \\
        & MHIM-MIL & 2.67 & \mstd{.6690}{.0358} & \mstd{.5874}{.0508} & \mstd{.8340}{.0128} & \mstd{.6970}{.0123} & \mstd{.6372}{.0158} & \mstd{.9155}{.0020} & \mstd{.8908}{.0183} & \mstd{.8703}{.0183} & \mstd{\best{.9759}}{.0030} & \mstd{.9465}{.0095} & \mstd{.9117}{.0155} & \mstd{.9769}{.0129} \\
        & DGR-MIL & 4.35 & \mstd{.7126}{.0363} & \mstd{.6684}{.0509} & \mstd{.8258}{.0386} & \mstd{.6964}{.0129} & \mstd{.6399}{.0178} & \mstd{.9043}{.0082} & \mstd{\second{.9036}}{.0225} & \mstd{.8829}{.0225} & \mstd{.9390}{.0224} & \mstd{.9455}{.0051} & \mstd{.9099}{.0098} & \mstd{.9724}{.0197} \\
        & 2DMambaMIL & 1.27 & \mstd{.7185}{.0290} & \mstd{.6710}{.0450} & \mstd{.8315}{.0310} & \mstd{.7012}{.0115} & \mstd{.6450}{.0160} & \mstd{.9105}{.0075} & \mstd{.9008}{.0190} & \mstd{\second{.8855}}{.0210} & \mstd{.9450}{.0180} & \mstd{\second{.9490}}{.0060} & \mstd{.9150}{.0110} & \mstd{.9755}{.0180} \\
        \midrule
        & \textbf{ResTopoMIL} & 1.15 & \mstd{\best{.7494}}{.0286} & \mstd{\best{.7142}}{.0420} & \mstd{\best{.9006}}{.0055} & \mstd{\best{.7546}}{.0094} & \mstd{\best{.7097}}{.0098} & \mstd{\best{.9426}}{.0010} & \mstd{\best{.9157}}{.0085} & \mstd{\best{.9135}}{.0087} & \mstd{\second{.9753}}{.0029} & \mstd{\best{.9568}}{.0095} & \mstd{\best{.9308}}{.0166} & \mstd{\best{.9838}}{.0049} \\
        \midrule
        \multirow{9}{*}{\textbf{CPath}} & AB-MIL & \best{.59} & \mstd{.7250}{.0210} & \mstd{.6415}{.0225} & \mstd{.8512}{.0105} & \mstd{.6854}{.0092} & \mstd{.6652}{.0085} & \mstd{.9185}{.0042} & \mstd{.8810}{.0075} & \mstd{.8510}{.0082} & \mstd{.9412}{.0085} & \mstd{.9250}{.0075} & \mstd{.9010}{.0115} & \mstd{.9605}{.0035} \\
        & CLAM-SB & \second{.79} & \mstd{.7185}{.0195} & \mstd{.6120}{.0215} & \mstd{\second{.8580}}{.0125} & \mstd{.6820}{.0088} & \mstd{.6610}{.0065} & \mstd{.9150}{.0038} & \mstd{.8855}{.0092} & \mstd{.8580}{.0095} & \mstd{.9485}{.0075} & \mstd{.9150}{.0085} & \mstd{.9085}{.0125} & \mstd{\second{.9685}}{.0045} \\
        & DS-MIL & 1.20 & \mstd{\second{.7305}}{.0285} & \mstd{\second{.6525}}{.0320} & \mstd{.8550}{.0152} & \mstd{.6815}{.0065} & \mstd{.6612}{.0052} & \mstd{\second{.9205}}{.0068} & \mstd{.8850}{.0115} & \mstd{.8530}{.0098} & \mstd{.9430}{.0092} & \mstd{.8910}{.0225} & \mstd{\second{.9110}}{.0135} & \mstd{.9622}{.0041} \\
        & TransMIL & 2.67 & \mstd{.6150}{.0185} & \mstd{.5255}{.0415} & \mstd{.8125}{.0128} & \mstd{.6855}{.0105} & \mstd{.6420}{.0150} & \mstd{.9024}{.0055} & \mstd{\second{.8950}}{.0145} & \mstd{.8655}{.0140} & \mstd{.9515}{.0075} & \mstd{.9250}{.0115} & \mstd{.8850}{.0175} & \mstd{.9645}{.0160} \\
        & ILRA-MIL & 3.68 & \mstd{.5900}{.0250} & \mstd{.5105}{.0385} & \mstd{.7850}{.0210} & \mstd{\second{.7255}}{.0080} & \mstd{.6710}{.0115} & \mstd{.9120}{.0045} & \mstd{.8820}{.0165} & \mstd{.8710}{.0162} & \mstd{.9485}{.0082} & \mstd{.9310}{.0065} & \mstd{.8950}{.0105} & \mstd{.9525}{.0115} \\
        & MHIM-MIL & 2.67 & \mstd{.6350}{.0380} & \mstd{.5510}{.0520} & \mstd{.8015}{.0165} & \mstd{.6750}{.0135} & \mstd{.6155}{.0185} & \mstd{.8985}{.0035} & \mstd{.8910}{.0195} & \mstd{.8550}{.0195} & \mstd{\second{.9610}}{.0048} & \mstd{.9310}{.0105} & \mstd{.8950}{.0165} & \mstd{.9658}{.0142} \\
        & DGR-MIL & 4.35 & \mstd{.6800}{.0350} & \mstd{.6350}{.0485} & \mstd{.7950}{.0305} & \mstd{.6720}{.0145} & \mstd{.6185}{.0195} & \mstd{.8845}{.0095} & \mstd{.8850}{.0235} & \mstd{.8650}{.0245} & \mstd{.9220}{.0250} & \mstd{.8950}{.0115} & \mstd{.8950}{.0155} & \mstd{.9605}{.0210} \\
        & 2DMambaMIL & 1.27 & \mstd{.6905}{.0285} & \mstd{.6420}{.0465} & \mstd{.8080}{.0255} & \mstd{.6255}{.0185} & \mstd{\second{.6815}}{.0125} & \mstd{.8912}{.0082} & \mstd{.8910}{.0185} & \mstd{\second{.8750}}{.0215} & \mstd{.9315}{.0195} & \mstd{\second{.9350}}{.0075} & \mstd{.9010}{.0125} & \mstd{.9640}{.0185} \\
        \midrule
        & \textbf{ResTopoMIL} & 1.15 & \mstd{\best{.7385}}{.0255} & \mstd{\best{.6952}}{.0435} & \mstd{\best{.8715}}{.0085} & \mstd{\best{.7350}}{.0110} & \mstd{\best{.6885}}{.0105} & \mstd{\best{.9258}}{.0025} & \mstd{\best{.9010}}{.0095} & \mstd{\best{.8955}}{.0095} & \mstd{\best{.9685}}{.0032} & \mstd{\best{.9420}}{.0105} & \mstd{\best{.9150}}{.0175} & \mstd{\best{.9750}}{.0052} \\
        \bottomrule
    \end{tabular}
    }

\end{table*}

With UNI, ResTopoMIL achieved the highest mean F1 on all four datasets. Its AUC margins over the strongest baseline were 0.0166 on BRACS, 0.0117 on PANDA and 0.0024 on BRCA. On NSCLC, MHIM-MIL had slightly higher AUC, 0.9759 versus 0.9753, while ResTopoMIL had higher accuracy and F1. The method's relative performance therefore depended on both task and metric.

With CTransPath, ResTopoMIL led all 12 reported dataset--metric combinations. AUC margins were 0.0135, 0.0053, 0.0075 and 0.0065 on BRACS, PANDA, NSCLC and BRCA, respectively. The model used 1.15 million trainable parameters, compared with 2.67 million for TransMIL and 4.35 million for DGR-MIL. Parameter counts describe model size and are not runtime measurements.

\subsection{Survival prediction and patient-level reliability}
The survival tasks evaluate transfer from diagnostic categories to time-to-event prediction. ResTopoMIL achieved the highest mean C-index on all five cohorts in Table~\ref{tab:result_survival_prediction}. Gains were clearest on LUAD and STAD; the KIRC margin over the strongest baseline was only 0.0002.

\begin{table}[!t]
    \centering
    \footnotesize

    \caption{Survival prediction results. Each cell reports mean$\pm$std concordance index over five random seeds on five TCGA cohorts. Higher is better. Bold and underlined means denote the best and second-best values per cohort, respectively.}
    \label{tab:result_survival_prediction}
    \setlength{\tabcolsep}{4.5pt}
    \renewcommand{\arraystretch}{1.02}
    \fitwidth{\linewidth}{
    \begin{tabular}{l c c c c c}
        \toprule
        \textbf{Method} & \textbf{KIRC} & \textbf{KIRP} & \textbf{LUAD} & \textbf{STAD} & \textbf{UCEC} \\
        \midrule
        AB-MIL &
        \mstd{0.5694}{.0279} & \mstd{0.7091}{.0592} & \mstd{0.5942}{.0120} & \mstd{0.5871}{.0090} & \mstd{0.6220}{.0241} \\
        CLAM-SB &
        \mstd{0.5705}{.0101} & \mstd{0.7034}{.0565} & \mstd{0.6192}{.0108} & \mstd{0.5829}{.0243} & \mstd{0.6034}{.0398} \\
        DS-MIL &
        \mstd{0.5832}{.0207} & \mstd{0.6875}{.0456} & \mstd{0.5669}{.0145} & \mstd{0.6090}{.0187} & \mstd{0.6368}{.0417} \\
        TransMIL &
        \mstd{0.5677}{.0162} & \mstd{0.7000}{.0238} & \mstd{0.6148}{.0124} & \mstd{0.6204}{.0359} & \mstd{0.6595}{.0486} \\
        ILRA-MIL &
        \mstd{0.7276}{.0271} & \mstd{0.7272}{.1322} & \mstd{0.6067}{.0156} & \mstd{0.6589}{.0354} & \mstd{0.6767}{.0308} \\
        MHIM-MIL &
        \mstd{0.7170}{.0143} & \mstd{\second{0.8073}}{.0545} & \mstd{0.5491}{.0205} & \mstd{0.6357}{.0321} & \mstd{0.6474}{.0704} \\
        DGR-MIL &
        \mstd{0.7298}{.0068} & \mstd{0.8071}{.0265} & \mstd{0.6245}{.0117} & \mstd{\second{0.6625}}{.0122} & \mstd{0.6976}{.0104} \\
        2DMambaMIL &
        \mstd{\second{0.7311}}{.0110} & \mstd{0.8027}{.0250} & \mstd{\second{0.6290}}{.0095} & \mstd{0.6515}{.0150} & \mstd{\second{0.7020}}{.0140} \\
        \midrule
        \textbf{ResTopoMIL} &
        \mstd{\best{0.7313}}{.0104} & \mstd{\best{0.8182}}{.0227} & \mstd{\best{0.6457}}{.0070} & \mstd{\best{0.6807}}{.0086} & \mstd{\best{0.7058}}{.0128} \\
        \bottomrule
    \end{tabular}
    }

\end{table}

Patient-cluster tests bounded the reliability of the primary predictive comparisons (Table~\ref{tab:paired_inference}). BRACS, PANDA, LUAD and STAD retained significance after Holm adjustment. The remaining positive estimates had confidence intervals containing zero, while NSCLC had a small negative estimate. Across classification AUC and survival C-index, mean primary performance improved on eight of nine endpoints, with statistically supported gains confined to four.

\begin{table}[!t]
    \centering
    \scriptsize
    \caption{Paired patient-cluster inference against the strongest original UNI baseline. $\Delta$ is ResTopoMIL minus comparator. Confidence intervals are marginal 95\% intervals; $p_H$ is Holm-adjusted across nine endpoints.}
    \label{tab:paired_inference}
    \setlength{\tabcolsep}{2.5pt}
    \fitwidth{\linewidth}{%
    \begin{tabular}{llccc}
        \toprule
        Endpoint & Comparator & $\Delta$ & 95\% CI & $p_H$ \\
        \midrule
        BRACS AUC & CLAM-SB & +0.0166 & [0.0082,0.0249] & 0.0025 \\
        PANDA AUC & DS-MIL & +0.0117 & [0.0063,0.0170] & 0.0016 \\
        NSCLC AUC & MHIM-MIL & -0.0006 & [-0.0024,0.0012] & 1.0000 \\
        BRCA AUC & CLAM-SB & +0.0024 & [-0.0015,0.0061] & 0.2582 \\
        KIRC C-index & 2DMambaMIL & +0.0002 & [-0.0062,0.0067] & 1.0000 \\
        KIRP C-index & MHIM-MIL & +0.0109 & [-0.0240,0.0460] & 1.0000 \\
        LUAD C-index & 2DMambaMIL & +0.0167 & [0.0113,0.0221] & $<$0.001 \\
        STAD C-index & DGR-MIL & +0.0182 & [0.0108,0.0255] & $<$0.001 \\
        UCEC C-index & 2DMambaMIL & +0.0038 & [-0.0049,0.0121] & 0.9580 \\
        \bottomrule
    \end{tabular}}
\end{table}

\subsection{Training schedule, residual capacity and design controls}
\begin{figure}[!t]
    \centering
    \includegraphics[width=\linewidth]{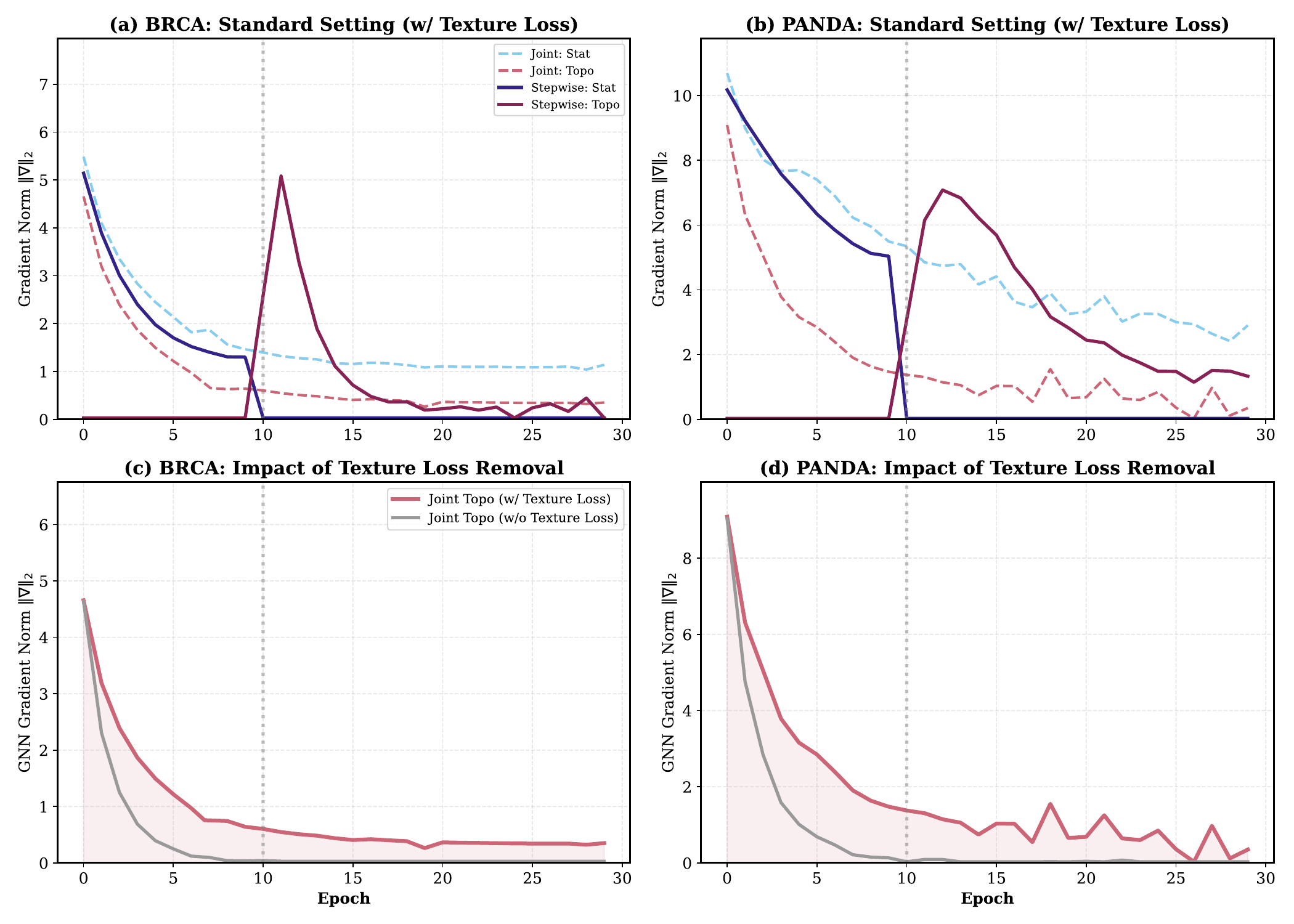}
    \caption{Gradient dynamics on BRCA and PANDA. Upper panels compare joint and staged optimization; the dashed vertical line marks the transition after ten warm-up epochs. The statistical stream is then frozen and graph training begins. Lower panels compare joint graph training with and without $\mathcal L_{texture}$. The plotted gradient norms are not separated into supervised and auxiliary components and provide descriptive optimization evidence.}
    \label{fig:gradient}
\end{figure}

The ablations test whether gains require the proposed training schedule and exceed added appearance capacity. Table~\ref{tab:ablation_main} contains 28 settings on PANDA and BRCA, including branch, fusion, graph and hyperparameter controls. Figure~\ref{fig:gradient} provides descriptive training traces for the optimization interpretation.

\paragraph{Training dynamics}
In the upper panels, staged training switched from the statistical stream to the graph stream after ten epochs. The graph gradient rose when graph optimization began, while the statistical gradient became zero because that stream was frozen. Joint-training graph gradients generally declined earlier and were weaker after this transition. The lower panels specifically compared joint graph training with and without $\mathcal L_{texture}$; removing the constraint produced faster gradient decay. These curves do not separate supervised and auxiliary gradients, so they do not identify the source of every gradient increase.

\begin{table*}[!t]
    \centering
    \footnotesize
    \caption{Consolidated ablation on PANDA and TCGA-BRCA with UNI features. Reported dispersion is standard deviation over five matched runs. All metrics are higher-is-better; bold and underlined values denote the highest and second-highest distinct reported means in each column, respectively, without implying significance. The default uses $K_{proto}=32$, $K_{knn}=8$, and $m=0.3$; each sweep changes only the named parameter.}
    \label{tab:ablation_main}
    \setlength{\tabcolsep}{3.0pt}
    \renewcommand{\arraystretch}{1.08}
    \fitwidth{\textwidth}{%
    \begin{tabular}{l ccc ccc}
        \toprule
        \multirow{2}{*}{\textbf{Variant}} & \multicolumn{3}{c}{\textbf{PANDA}} & \multicolumn{3}{c}{\textbf{TCGA-BRCA}} \\
        \cmidrule(lr){2-4}\cmidrule(lr){5-7}
        & Accuracy & F1 & AUC & Accuracy & F1 & AUC \\
        \midrule
        \multicolumn{7}{l}{\textit{Reference model}} \\
        \textbf{ResTopoMIL (default)} & \mstd{\second{0.7546}}{0.0094} & \mstd{\second{0.7097}}{0.0098} & \mstd{\best{0.9426}}{0.0010} & \mstd{\best{0.9568}}{0.0095} & \mstd{0.9308}{0.0166} & \mstd{0.9838}{0.0049} \\
        \midrule
        \multicolumn{7}{l}{\textit{A. Branches and optimization}} \\
        Statistical only & \mstd{0.6761}{0.0062} & \mstd{0.6113}{0.0075} & \mstd{0.9027}{0.0051} & \mstd{0.9138}{0.0069} & \mstd{0.8896}{0.0112} & \mstd{0.9486}{0.0035} \\
        Topological only & \mstd{0.7042}{0.0072} & \mstd{0.6598}{0.0085} & \mstd{0.9215}{0.0055} & \mstd{0.9382}{0.0068} & \mstd{0.8945}{0.0102} & \mstd{0.9608}{0.0048} \\
        Joint optimization & \mstd{0.7397}{0.0055} & \mstd{0.7025}{0.0065} & \mstd{0.9299}{0.0052} & \mstd{0.9432}{0.0055} & \mstd{\best{0.9318}}{0.0061} & \mstd{0.9773}{0.0022} \\
        Joint + statistical dropout & \mstd{0.7415}{0.0070} & \mstd{0.7038}{0.0075} & \mstd{0.9325}{0.0065} & \mstd{0.9448}{0.0048} & \mstd{0.9302}{0.0068} & \mstd{0.9780}{0.0030} \\
        Joint + curriculum schedule & \mstd{0.7442}{0.0052} & \mstd{0.7055}{0.0058} & \mstd{0.9348}{0.0045} & \mstd{0.9465}{0.0045} & \mstd{\second{0.9315}}{0.0055} & \mstd{0.9785}{0.0018} \\
        Joint + hard-instance mining & \mstd{0.7485}{0.0062} & \mstd{0.7091}{0.0065} & \mstd{0.9372}{0.0050} & \mstd{0.9492}{0.0052} & \mstd{0.9288}{0.0065} & \mstd{0.9792}{0.0021} \\
        Joint optimization (multi-LR) & \mstd{0.7468}{0.0051} & \mstd{0.7084}{0.0045} & \mstd{0.9352}{0.0048} & \mstd{0.9475}{0.0042} & \mstd{0.9235}{0.0054} & \mstd{0.9786}{0.0018} \\
        \midrule
        \multicolumn{7}{l}{\textit{B. Fusion alternatives}} \\
        Gated fusion & \mstd{0.7389}{0.0068} & \mstd{0.6991}{0.0075} & \mstd{0.9225}{0.0062} & \mstd{0.9432}{0.0051} & \mstd{0.9068}{0.0065} & \mstd{0.9802}{0.0042} \\
        Mixture of experts & \mstd{0.7471}{0.0052} & \mstd{0.6918}{0.0048} & \mstd{0.9305}{0.0035} & \mstd{0.9432}{0.0045} & \mstd{0.9108}{0.0055} & \mstd{\second{0.9897}}{0.0028} \\
        Concat + MLP & \mstd{0.7398}{0.0069} & \mstd{0.6928}{0.0074} & \mstd{0.9276}{0.0058} & \mstd{0.9302}{0.0057} & \mstd{0.9098}{0.0083} & \mstd{0.9862}{0.0031} \\
        Independent classifiers & \mstd{0.7352}{0.0085} & \mstd{0.6845}{0.0081} & \mstd{0.9256}{0.0062} & \mstd{0.9465}{0.0058} & \mstd{0.9112}{0.0095} & \mstd{0.9845}{0.0025} \\
        \midrule
        \multicolumn{7}{l}{\textit{C. Graph and prototype controls}} \\
        GAT backbone & \mstd{0.7489}{0.0062} & \mstd{0.7074}{0.0058} & \mstd{\second{0.9419}}{0.0041} & \mstd{0.9432}{0.0052} & \mstd{0.9068}{0.0061} & \mstd{0.9889}{0.0025} \\
        Random graph (training control) & \mstd{0.6815}{0.0148} & \mstd{0.5362}{0.0260} & \mstd{0.8286}{0.0150} & \mstd{0.8523}{0.0180} & \mstd{0.6749}{0.0250} & \mstd{0.8563}{0.0110} \\
        Fixed prototypes & \mstd{0.7335}{0.0065} & \mstd{0.6781}{0.0072} & \mstd{0.9349}{0.0055} & \mstd{0.9446}{0.0056} & \mstd{0.9098}{0.0094} & \mstd{0.9746}{0.0035} \\
        \midrule
        \multicolumn{7}{l}{\textit{D. Coordinate-specific constraint}} \\
        w/o $\mathcal{L}_{texture}$ & \mstd{0.7155}{0.0072} & \mstd{0.6970}{0.0081} & \mstd{0.9147}{0.0065} & \mstd{0.9418}{0.0054} & \mstd{0.9065}{0.0098} & \mstd{0.9762}{0.0038} \\
        \midrule
        \multicolumn{7}{l}{\textit{E. Prototype count}} \\
        $K_{proto}=8$ & \mstd{0.7405}{0.0071} & \mstd{0.6912}{0.0063} & \mstd{0.9285}{0.0051} & \mstd{0.9482}{0.0055} & \mstd{0.9145}{0.0089} & \mstd{0.9852}{0.0021} \\
        $K_{proto}=16$ & \mstd{0.7512}{0.0058} & \mstd{0.7065}{0.0042} & \mstd{0.9392}{0.0034} & \mstd{0.9515}{0.0038} & \mstd{0.9224}{0.0061} & \mstd{0.9889}{0.0015} \\
        $K_{proto}=64$ & \mstd{0.7544}{0.0052} & \mstd{\best{0.7098}}{0.0048} & \mstd{0.9415}{0.0031} & \mstd{0.9528}{0.0042} & \mstd{0.9245}{0.0065} & \mstd{0.9892}{0.0018} \\
        \midrule
        \multicolumn{7}{l}{\textit{F. Graph degree}} \\
        $K_{knn}=4$ & \mstd{0.7438}{0.0062} & \mstd{0.6955}{0.0059} & \mstd{0.9312}{0.0045} & \mstd{0.9502}{0.0049} & \mstd{0.9192}{0.0071} & \mstd{0.9875}{0.0019} \\
        $K_{knn}=16$ & \mstd{0.7495}{0.0055} & \mstd{0.7024}{0.0051} & \mstd{0.9368}{0.0037} & \mstd{0.9518}{0.0044} & \mstd{0.9218}{0.0068} & \mstd{0.9882}{0.0022} \\
        \midrule
        \multicolumn{7}{l}{\textit{G. Shuffle margin}} \\
        Margin $m=0.1$ & \mstd{0.7462}{0.0060} & \mstd{0.6988}{0.0055} & \mstd{0.9325}{0.0048} & \mstd{0.9510}{0.0040} & \mstd{0.9205}{0.0062} & \mstd{0.9880}{0.0018} \\
        Margin $m=0.2$ & \mstd{0.7535}{0.0048} & \mstd{0.7082}{0.0041} & \mstd{0.9408}{0.0035} & \mstd{0.9532}{0.0035} & \mstd{0.9248}{0.0055} & \mstd{0.9896}{0.0014} \\
        Margin $m=0.4$ & \mstd{\best{0.7548}}{0.0049} & \mstd{0.7094}{0.0042} & \mstd{0.9417}{0.0033} & \mstd{\second{0.9536}}{0.0036} & \mstd{0.9251}{0.0053} & \mstd{\best{0.9898}}{0.0014} \\
        Margin $m=0.5$ & \mstd{0.7505}{0.0052} & \mstd{0.7042}{0.0049} & \mstd{0.9385}{0.0039} & \mstd{0.9525}{0.0038} & \mstd{0.9230}{0.0058} & \mstd{0.9890}{0.0016} \\
        \midrule
        \multicolumn{7}{l}{\textit{H. Coordinate-free residuals and inference corruption}} \\
        Parameter-matched DeepSets residual & \mstd{0.6904}{0.0073} & \mstd{0.6280}{0.0086} & \mstd{0.9113}{0.0043} & \mstd{0.9234}{0.0065} & \mstd{0.9007}{0.0095} & \mstd{0.9597}{0.0032} \\
        AB-MIL residual & \mstd{0.7385}{0.0061} & \mstd{0.6931}{0.0070} & \mstd{0.9289}{0.0032} & \mstd{0.9472}{0.0058} & \mstd{0.9184}{0.0085} & \mstd{0.9752}{0.0027} \\
        Default, shuffled graph at inference & \mstd{0.6829}{0.0060} & \mstd{0.6168}{0.0071} & \mstd{0.9052}{0.0029} & \mstd{0.9155}{0.0062} & \mstd{0.8919}{0.0089} & \mstd{0.9509}{0.0037} \\
        \bottomrule
    \end{tabular}}
\end{table*}

\paragraph{Schedule beyond branch capacity}
The staged model exceeded joint optimization in AUC by 0.0127 on PANDA and 0.0065 on BRCA. Neither branch alone matched the complete model's AUC. Statistical dropout, curriculum scheduling, hard-instance mining and a tenfold graph learning rate narrowed the gap. Hard-instance mining was strongest among these alternatives, reaching 0.9372/0.9792 versus 0.9426/0.9838 for the default. Joint training nevertheless gave slightly higher BRCA F1, 0.9318 versus 0.9308. These comparisons support the schedule's predictive contribution without establishing superiority on every metric.

\paragraph{Appearance capacity beyond the anchor}
Coordinate-free residuals improved prediction even without physical neighborhoods (Panel H). They shared anchor logits, UNI features, five patient-level splits, optimizer family and Stage-2 budget. A parameter-matched two-layer node MLP with mean pooling increased PANDA/BRCA AUC by 0.0086/0.0111 over the anchor. An AB-MIL residual reached 0.9289/0.9752, remaining 0.0137/0.0086 below the full model. The anchor therefore left usable appearance information, but these controls did not recover all graph-model gains.

\paragraph{Fusion and graph design}
Additive prediction gave higher PANDA AUC and higher accuracy on both datasets than the tested fusion alternatives. However, MoE, independent classifiers and Concat + MLP achieved higher BRCA AUCs, respectively 0.9897, 0.9845 and 0.9862 versus 0.9838. Replacing GCN with GAT gave similar PANDA AUC and higher BRCA AUC, 0.9419/0.9889. Training on random graphs reduced AUC to 0.8286/0.8563, while fixing the prototypes also reduced performance. The intervention was not restricted to GCN, and its default fusion was not universally optimal.

\paragraph{Coordinate-specific constraint and sensitivity}
Removing $\mathcal L_{texture}$ reduced PANDA/BRCA AUC by 0.0279/0.0076 (Panel D). Across the prototype-count, neighbor-count and margin sweeps, PANDA AUC ranged over 0.9285--0.9426, 0.9312--0.9426 and 0.9325--0.9426. The defaults were not uniformly best: margin 0.4 gave BRCA AUC 0.9898, compared with 0.9838. The sweeps varied one parameter at a time and did not test interactions or loss-weight sensitivity.

\subsection{Prediction depends on the supplied neighborhoods}
\label{sec:spatial_evidence}
\begin{figure}[!t]
    \centering
    \includegraphics[width=\linewidth]{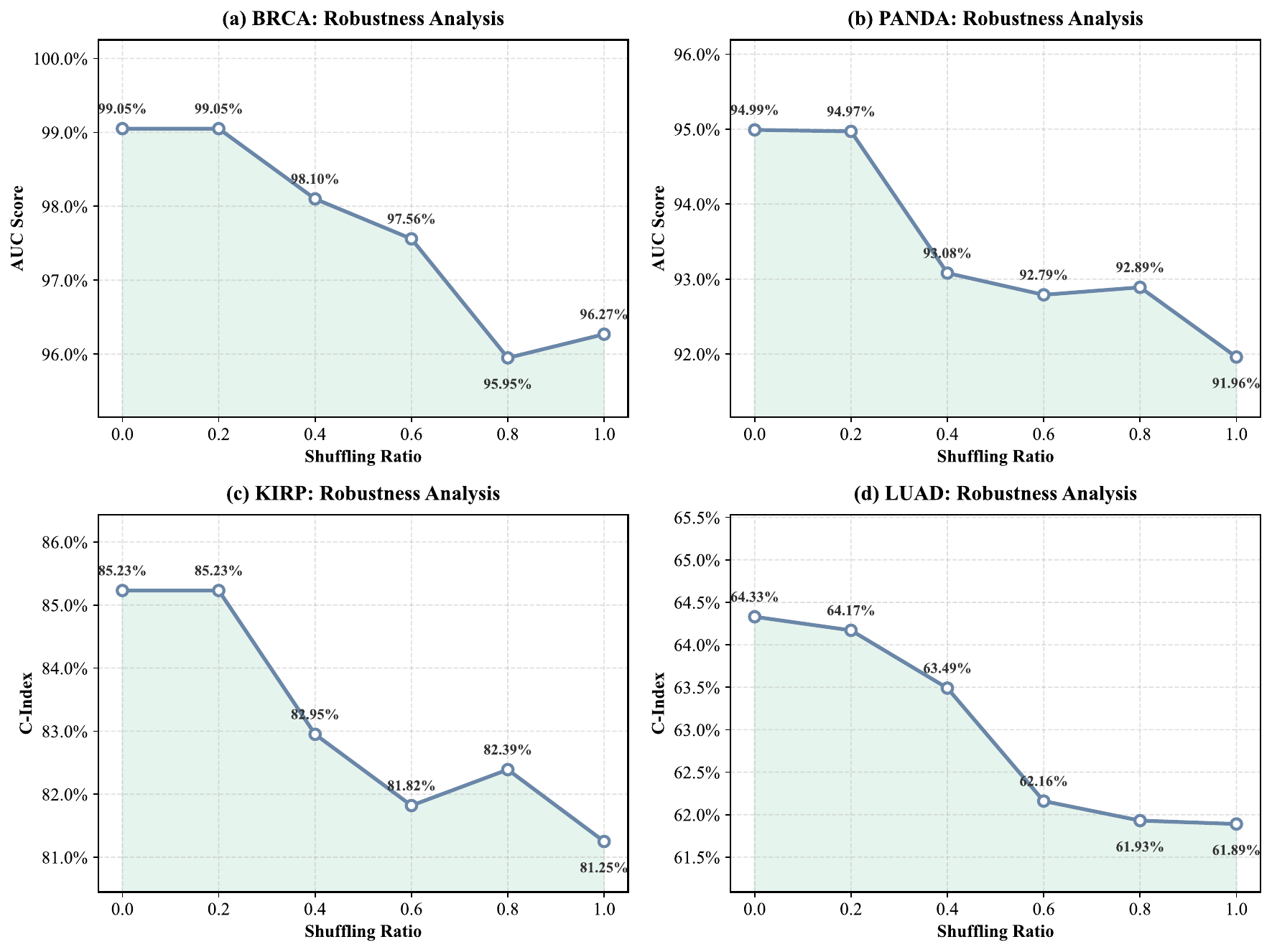}
    \caption{Progressive coordinate shuffling with fixed patch embeddings. Panels report AUC on TCGA-BRCA/PANDA and C-index on TCGA-KIRP/TCGA-LUAD. Curves are representative-seed trajectories, with no uncertainty bands; they are not five-seed aggregate estimates.}
    \label{fig:shuffling_main}
\end{figure}

Coordinate perturbations test whether the trained predictor uses spatial inputs. Patch embeddings remain fixed, coordinate assignments are permuted, and coordinate-derived inputs are rebuilt without retraining. Figure~\ref{fig:shuffling_main} reports progressive perturbations, while Table~\ref{tab:shuffle_endpoint_main} retains endpoint comparisons for 2DMambaMIL and PatchGCN.

\begin{table*}[!t]
    \centering
    \footnotesize
    \caption{Spatial-baseline AUC before and after complete coordinate reassignment with fixed patch embeddings. Clean and Shuf. denote original and shuffled endpoints; $\Delta=\mathrm{Shuf.}-\mathrm{Clean}$. Reported endpoints have no run-level dispersion and are not five-seed aggregate estimates.}
    \label{tab:shuffle_endpoint_main}
    \label{tab:shuffle_reference_main}
    \setlength{\tabcolsep}{3pt}
    \renewcommand{\arraystretch}{1.05}
    \fitwidth{\textwidth}{%
    \begin{tabular}{lccc ccc ccc ccc}
        \toprule
        \multirow{2}{*}{\textbf{Method}} & \multicolumn{3}{c}{\textbf{BRACS}} & \multicolumn{3}{c}{\textbf{PANDA}} & \multicolumn{3}{c}{\textbf{NSCLC}} & \multicolumn{3}{c}{\textbf{BRCA}} \\
        \cmidrule(lr){2-4} \cmidrule(lr){5-7} \cmidrule(lr){8-10} \cmidrule(lr){11-13}
        & Clean & Shuf. & $\Delta$ & Clean & Shuf. & $\Delta$ & Clean & Shuf. & $\Delta$ & Clean & Shuf. & $\Delta$ \\
        \midrule
        2DMambaMIL & 0.8315 & 0.8237 & -0.0078 & 0.9105 & 0.9032 & -0.0073 & 0.9450 & 0.9418 & -0.0032 & 0.9755 & 0.9709 & -0.0046 \\
        PatchGCN & 0.8738 & 0.8520 & -0.0218 & 0.9274 & 0.9127 & -0.0147 & 0.9711 & 0.9642 & -0.0069 & 0.9801 & 0.9700 & -0.0101 \\
        \bottomrule
    \end{tabular}}
    \par\smallskip
    \begin{minipage}{\textwidth}
    \footnotesize
    PatchGCN rebuilds its graph; 2DMambaMIL rebuilds its coordinate-derived 2D sequence. $\Delta$ measures sensitivity, not a performance ranking.
    \end{minipage}
\end{table*}

Performance generally declined, with local fluctuations, as shuffling increased on BRCA, PANDA, KIRP and LUAD. On PANDA, complete reassignment reduced 2DMambaMIL/PatchGCN AUC by 0.0073/0.0147, compared with 0.0374 for ResTopoMIL under identical embeddings and permutations. Mean absolute probability changes were 0.0090/0.0170 versus 0.0390; label-flip rates were 1.8\%/3.2\% versus 7.8\%. Larger responses indicated stronger dependence on the supplied arrangement, rather than a ranking of clinical reasoning.

Five-run inference corruption confirmed the contrast with coordinate-free residual capacity. Shuffling the full model's graph reduced PANDA/BRCA AUC to 0.9052/0.9509 (Table~\ref{tab:ablation_main}, Panel H), close to the anchor's 0.9027/0.9486. Coordinate-free controls in Figure~\ref{fig:teaser}(c) were nearly invariant because their effective inputs were unchanged. TransMIL's padded positional grid also need not change under physical-coordinate reassignment \citep{shao2021transmil}.

The composition-only toy task provided a negative control for unnecessary spatial dependence. ResTopoMIL's clean/shuffled Dataset A AUC was 0.9940/0.9938. Its mean absolute probability change was 0.0012, with 0.2\% label flips. The model thus showed little perturbation response in this controlled setting without a spatial label signal.

\subsection{Neighborhood-preserving and neighborhood-damaging controls}
\label{sec:extended_controls}
Preserving neighborhoods left PANDA performance nearly unchanged, while graph damage reduced it (Table~\ref{tab:graph_controls_main}). Translation, rotation, reflection and uniform scaling retained AUC 0.9426. Jitter at $0.10\times d_{NN}$ retained 96.1\% of edges and produced AUC 0.9421. Here $d_{NN}$ denotes the nearest-neighbor distance used to scale coordinate jitter.

\begin{table}[!t]
\centering\footnotesize
\caption{PANDA neighborhood controls. Edge retention refers to original KNN edges; $\Delta$ is perturbed minus clean AUC. Only AUC is ranked, with the best and second-best distinct values bold and underlined; retention and $\Delta$ are diagnostics.}
\label{tab:graph_controls_main}
\setlength{\tabcolsep}{3pt}
\fitwidth{\linewidth}{%
\begin{tabular}{lccc}
\toprule
Perturbation & Retention & AUC & $\Delta$ \\
\midrule
None & 100\% & \best{0.9426} & 0.0000 \\
Translation/rotation/reflection/scaling & 100\% & \best{0.9426} & 0.0000 \\
Jitter ($0.05\times d_{NN}$) & 98.8\% & \second{0.9425} & -0.0001 \\
Jitter ($0.10\times d_{NN}$) & 96.1\% & 0.9421 & -0.0005 \\
Degree-preserving rewiring, 25\% & $\approx$75\% & 0.9349 & -0.0077 \\
Degree-preserving rewiring, 50\% & $\approx$50\% & 0.9237 & -0.0189 \\
Degree-preserving rewiring, 100\% & $\approx$0\% & 0.9070 & -0.0356 \\
Complete coordinate permutation & $\approx$0\% & 0.9052 & -0.0374 \\
\bottomrule
\end{tabular}}
\end{table}

Connectivity- and degree-preserving double-edge swaps within original components reduced AUC to 0.9349, 0.9237 and 0.9070 at 25\%, 50\% and 100\% rewiring. Affected-node counts were controlled, with approximately matched edge-length and node-feature-similarity distributions. These controls associated prediction changes with neighborhood damage, while allowing higher-order or non-pathological correlations to contribute. Run-level dispersion was unavailable for these endpoints.

\subsection{Localization and anatomy-stratified perturbations}
CAMELYON-16 tested whether learned patch scores aligned with annotated anatomy \citep{ehteshami2017diagnostic}. Classifier training used slide labels alone. Official tumor masks were used for localization evaluation, validation-threshold selection and post-hoc perturbation stratification, without entering classifier inputs.

\paragraph{Patch scoring and localization protocol}
Evaluation used $256\times256$ patches at $20\times$, excluding patches outside the Otsu tissue mask. A patch was positive when at least 1\% of its area overlapped the tumor mask. ResTopoMIL scored each patch using its pre-pooling target-class residual logit, $s_i=\mathbf W_{topo,y}^{\top}\mathbf H_i^{(2)}$. Baselines used their native attention or instance scores. All scores were min--max normalized within each slide, without smoothing, morphological refinement or a tumor-shape prior.

Each method's Dice threshold maximized mean validation Dice and was then fixed for testing. Connected components in thresholded patch grids defined FROC candidates. A candidate was a true positive if its maximum-score patch lay inside an annotated tumor. The reported FROC summary averaged seven operating points: $1/8$, $1/4$, $1/2$, $1$, $2$, $4$ and $8$ false positives per slide. This patch-grid summary differs from the official six-point CAMELYON-16 score, which excludes $1/8$ \citep{ehteshami2017diagnostic}.

\begin{table}[!t]
    \centering
    \footnotesize
    \caption{CAMELYON-16 localization (upper panel) and post-hoc anatomy-stratified graph perturbation (lower panel). FROC uses the reported seven-point patch-grid summary. Masks stratify perturbations after training; higher values are better. Best and second-best values are bold and underlined within each panel; ties share formatting.}
    \label{tab:camelyon_main}
    \setlength{\tabcolsep}{3pt}
    \begin{tabular}{lccc}
        \toprule
        \textbf{Method} & \textbf{Dice} & \textbf{Specificity} & \textbf{FROC} \\
        \midrule
        AB-MIL & 0.412 & 0.985 & 0.3952 \\
        CLAM-SB & 0.459 & 0.987 & 0.4257 \\
        DS-MIL & 0.259 & 0.863 & 0.4506 \\
        TransMIL & 0.103 & \best{0.999} & \second{0.4866} \\
        DTFD-MIL & 0.525 & \best{0.999} & 0.4712 \\
        MHIM-MIL & \second{0.548} & 0.992 & 0.4815 \\
        2DMambaMIL & 0.475 & \second{0.995} & 0.4520 \\
        \textbf{ResTopoMIL} & \best{0.624} & \best{0.999} & \best{0.5483} \\
        \midrule
        \textbf{Graph condition} & \textbf{Slide AUC} & \textbf{Dice} & \textbf{FROC} \\
        Original & \best{0.973} & \best{0.624} & \best{0.5483} \\
        Random-edge rewiring & 0.944 & 0.581 & 0.5210 \\
        Normal-interior rewiring & \second{0.950} & \second{0.593} & \second{0.5290} \\
        Tumor-interior rewiring & 0.941 & 0.570 & 0.5120 \\
        Tumor--normal interface & 0.921 & 0.522 & 0.4810 \\
        \bottomrule
    \end{tabular}
\end{table}

\begin{figure*}[!t]
    \centering
    % Internal figure labels; retain the known source identifiers for traceability.
    % Case-01 -> Figure/attnheap.pdf.
    % Case-02 -> TCGA-A2-A3XY; Case-03 -> TCGA-B6-A0IA; Case-04 -> TCGA-LL-A442.
    \subcaptionbox{Case-01}{\includegraphics[width=0.48\textwidth]{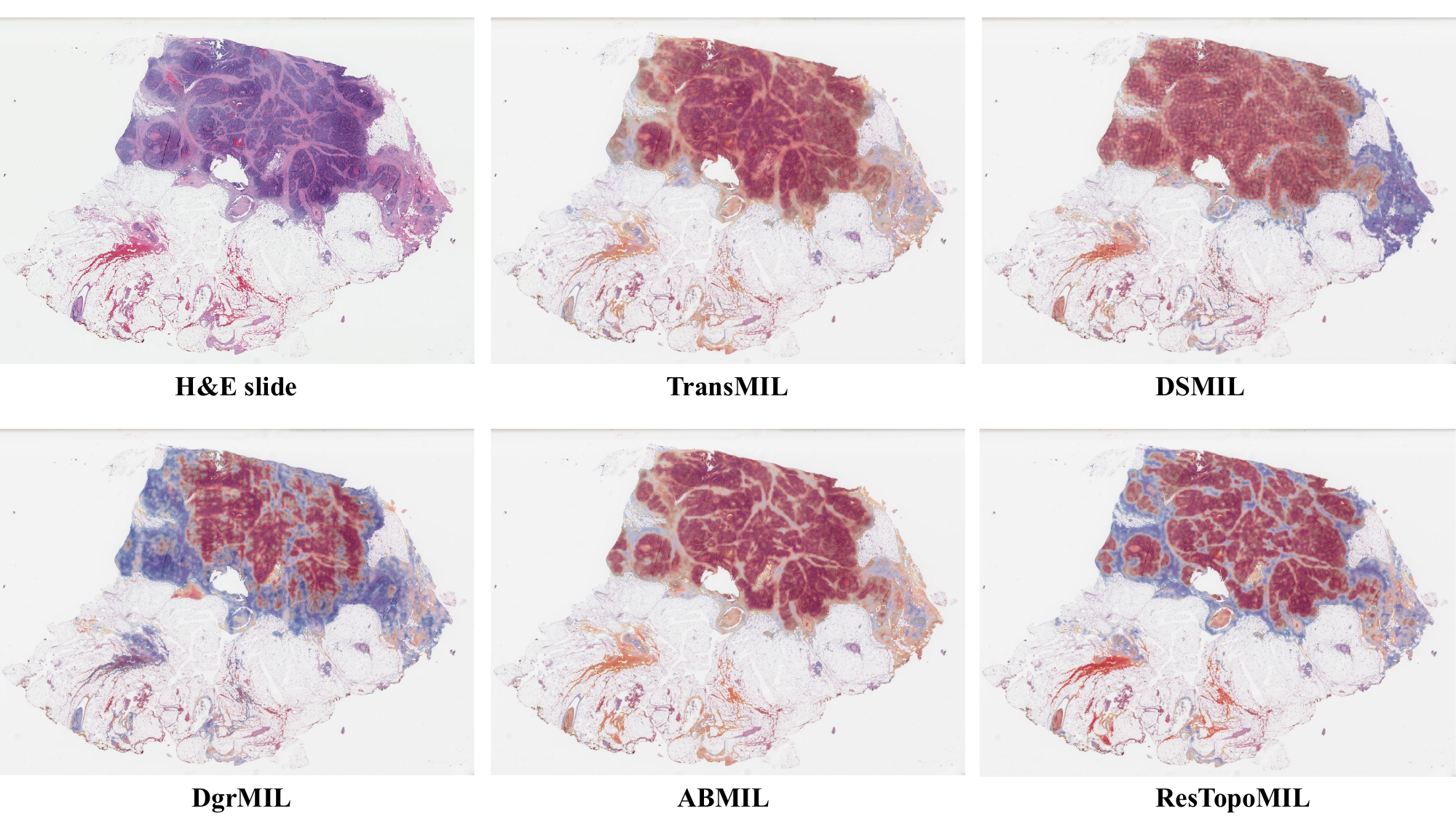}}\hfill
    \subcaptionbox{Case-02}{\includegraphics[width=0.48\textwidth]{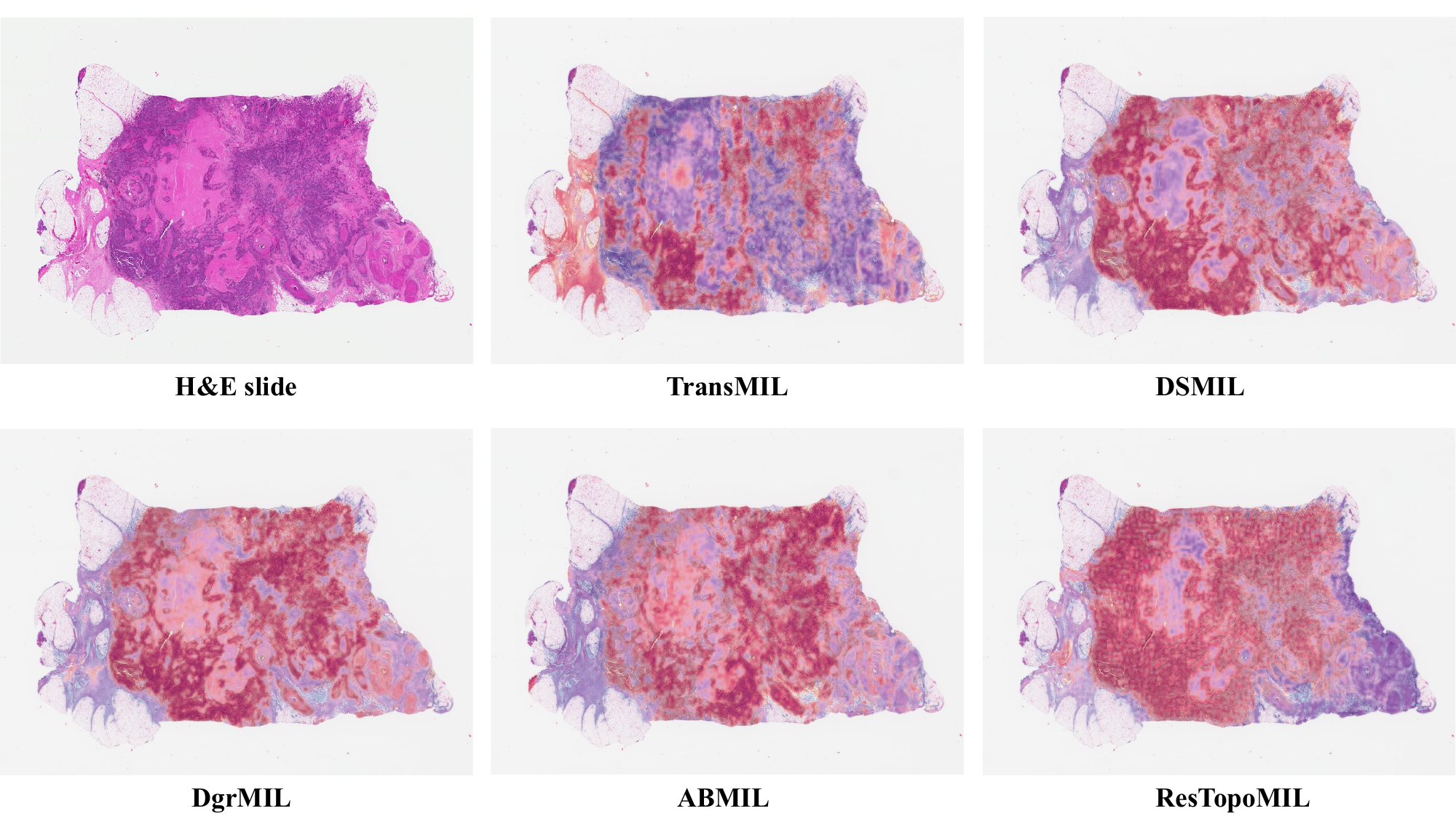}}\\[4pt]
    \subcaptionbox{Case-03}{\includegraphics[width=0.48\textwidth]{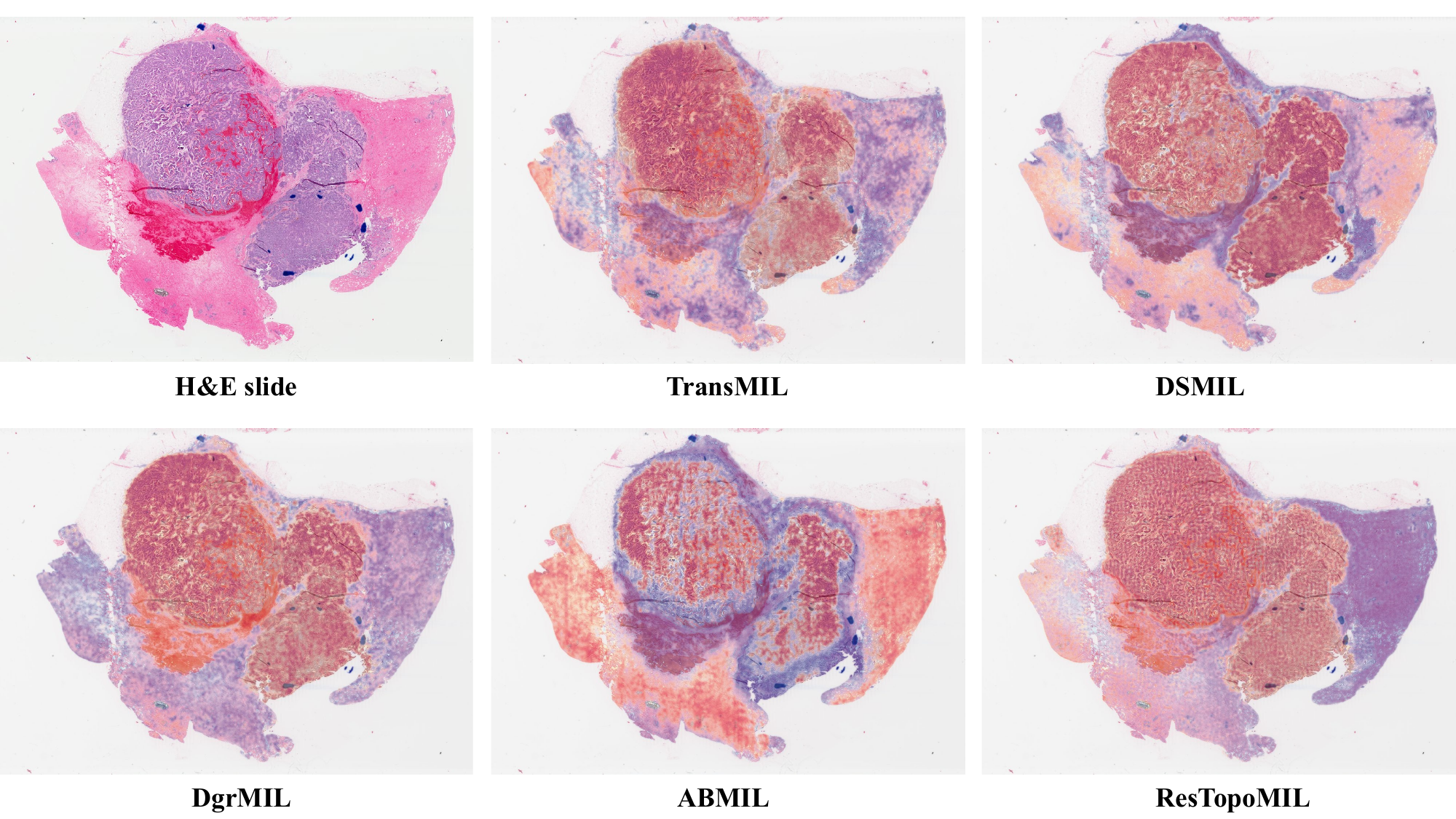}}\hfill
    \subcaptionbox{Case-04}{\includegraphics[width=0.48\textwidth]{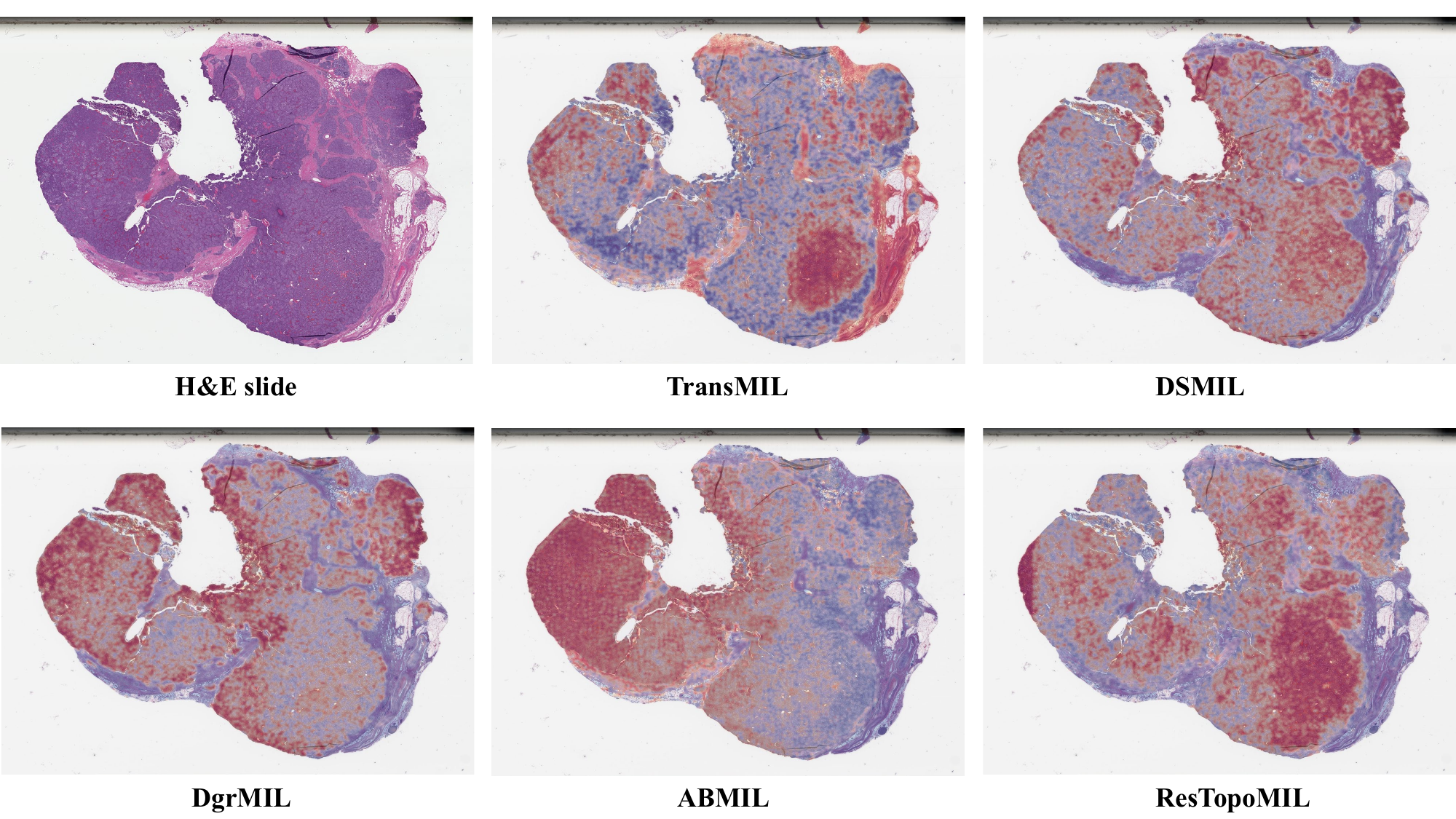}}
    \caption{Qualitative H\&E and native heatmap comparisons across five MIL methods. Warmer colors indicate higher within-slide normalized scores. Non-tissue patches are masked, with no smoothing, morphological refinement, or tumor-shape prior. The examples illustrate contiguity and background leakage, rather than segmentation accuracy. Case labels are internal to this figure.}
    \label{fig:heatmap_main}
\end{figure*}

\paragraph{Localization results}
ResTopoMIL exceeded the strongest baseline by 0.076 Dice and 0.0617 FROC (Table~\ref{tab:camelyon_main}). TransMIL and DTFD-MIL matched its specificity of 0.999, but had lower Dice. These comparisons measured patch-grid lesion coverage and background suppression; run-level uncertainty was unavailable. Figure~\ref{fig:heatmap_main} illustrates native heatmaps for four selected cases, including tumor-core concentration, boundary continuity, invasive-front patterns and background leakage. The cases complement mask-based metrics without estimating cohort-wide visual reliability.

\paragraph{Anatomical perturbation and score relevance}
Post-hoc tumor masks stratified connectivity- and degree-preserving graph rewiring. Rewiring tumor--normal interfaces produced the largest AUC, Dice and FROC losses among the reported conditions (Table~\ref{tab:camelyon_main}, lower panel). Removing the highest-scoring 10\% of patches reduced slide AUC from 0.973 to 0.835, compared with 0.961 after matched-size random removal. Retaining only those patches gave AUC 0.903. These checks linked scores and neighborhood dependence to annotated anatomy, without identifying a unique causal mechanism or providing a full deletion curve.

\subsection{Anchor convergence and patch-scale robustness}
Longer warm-up did not materially alter the composition reference in the tested histogram family. Extending Stage 1 from ten to twenty epochs or using early stopping changed anchor AUC by at most 0.0002 and final AUC by 0.0001 (Table~\ref{tab:warmup_main}). This comparison addressed premature freezing within the chosen anchor, while the coordinate-free residuals showed that the anchor still omitted appearance information.

\begin{table}[!t]
\centering\footnotesize
\caption{Warm-up schedule analysis. Cells give reported AUC without run-level dispersion. Stat./Full denote the anchor and complete model. Bold and underlined AUCs mark the best and second-best distinct values per column; ties share formatting.}
\label{tab:warmup_main}
\setlength{\tabcolsep}{4pt}
\begin{tabular}{lcccc}
\toprule
& \multicolumn{2}{c}{PANDA} & \multicolumn{2}{c}{BRCA} \\
Schedule & Stat. & Full & Stat. & Full \\
\midrule
10 epochs & 0.9027 & \second{0.9426} & 0.9486 & \second{0.9838} \\
20 epochs & \best{0.9029} & 0.9425 & \best{0.9488} & \best{0.9839} \\
Early stopping & \second{0.9028} & \best{0.9427} & \second{0.9487} & \second{0.9838} \\
\bottomrule
\end{tabular}
\end{table}

Across five matched runs, the anchor retained 91.0\%$\pm$1.2\% and 92.7\%$\pm$1.0\% of full-model AUC improvement over a class-prior predictor on PANDA and BRCA. The ratio was $(\mathrm{AUC}_{stat}-.5)/(\mathrm{AUC}_{full}-.5)$. The corresponding cross-entropy fraction, $(\mathrm{CE}_{prior}-\mathrm{CE}_{stat})/(\mathrm{CE}_{prior}-\mathrm{CE}_{full})$, was 86.9\%$\pm$2.0\% and 91.4\%$\pm$1.6\%. These ratios describe normalized predictive improvements rather than a decomposition of label information.

Matched patch-scale comparisons retained the relative advantage over 2DMambaMIL on BRACS (Table~\ref{tab:patch_scale_main}). Patient-level splits, UNI extraction, training procedures and five seeds were shared. ResTopoMIL exceeded the comparator on all three mean metrics at both patch sizes. With $512\times512$ patches, its accuracy, F1 and AUC margins were 0.0250, 0.0350 and 0.0220. Both methods gained accuracy with larger patches, but field of view, instance count and graph granularity changed together.

\begin{table}[!t]
\centering\footnotesize
\caption{Three-class BRACS under matched patch-scale protocols. Entries are mean$\pm$standard deviation over five runs. Best and second-best means are bold and underlined within each patch size.}
\label{tab:patch_scale_main}
\setlength{\tabcolsep}{2.5pt}
\fitwidth{\linewidth}{%
\begin{tabular}{llccc}
\toprule
Patch & Method & Accuracy & F1 & AUC \\
\midrule
$256^2$ & 2DMambaMIL & \mstd{\second{0.7185}}{0.0290} & \mstd{\second{0.6710}}{0.0450} & \mstd{\second{0.8315}}{0.0310} \\
& ResTopoMIL & \mstd{\best{0.7494}}{0.0286} & \mstd{\best{0.7142}}{0.0420} & \mstd{\best{0.9006}}{0.0055} \\
$512^2$ & 2DMambaMIL & \mstd{\second{0.7420}}{0.0250} & \mstd{\second{0.6860}}{0.0280} & \mstd{\second{0.8840}}{0.0210} \\
& ResTopoMIL & \mstd{\best{0.7670}}{0.0220} & \mstd{\best{0.7210}}{0.0250} & \mstd{\best{0.9060}}{0.0180} \\
\bottomrule
\end{tabular}}
\end{table}

\subsection{Representation diagnostics and computational scope}
\begin{figure}[!t]
    \centering
    \includegraphics[width=\linewidth]{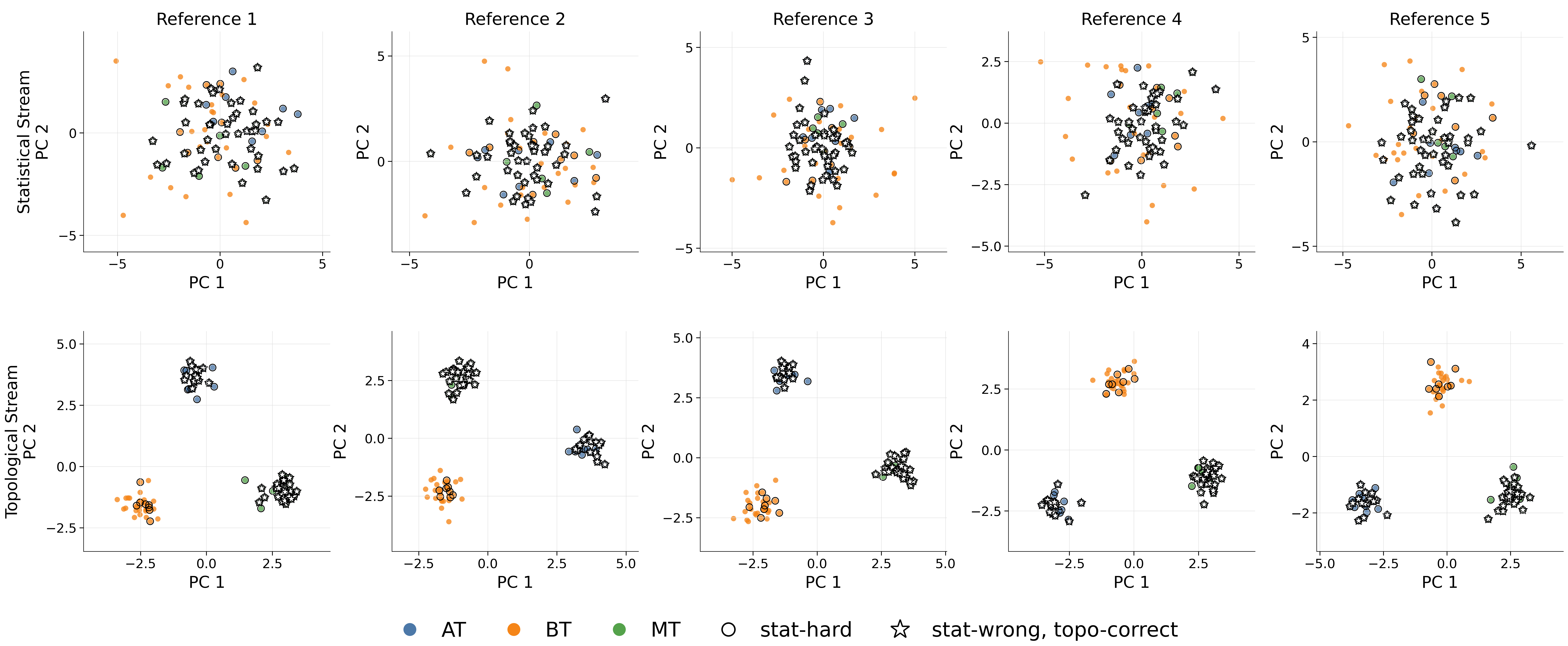}\\[-1pt]
    \includegraphics[width=\linewidth]{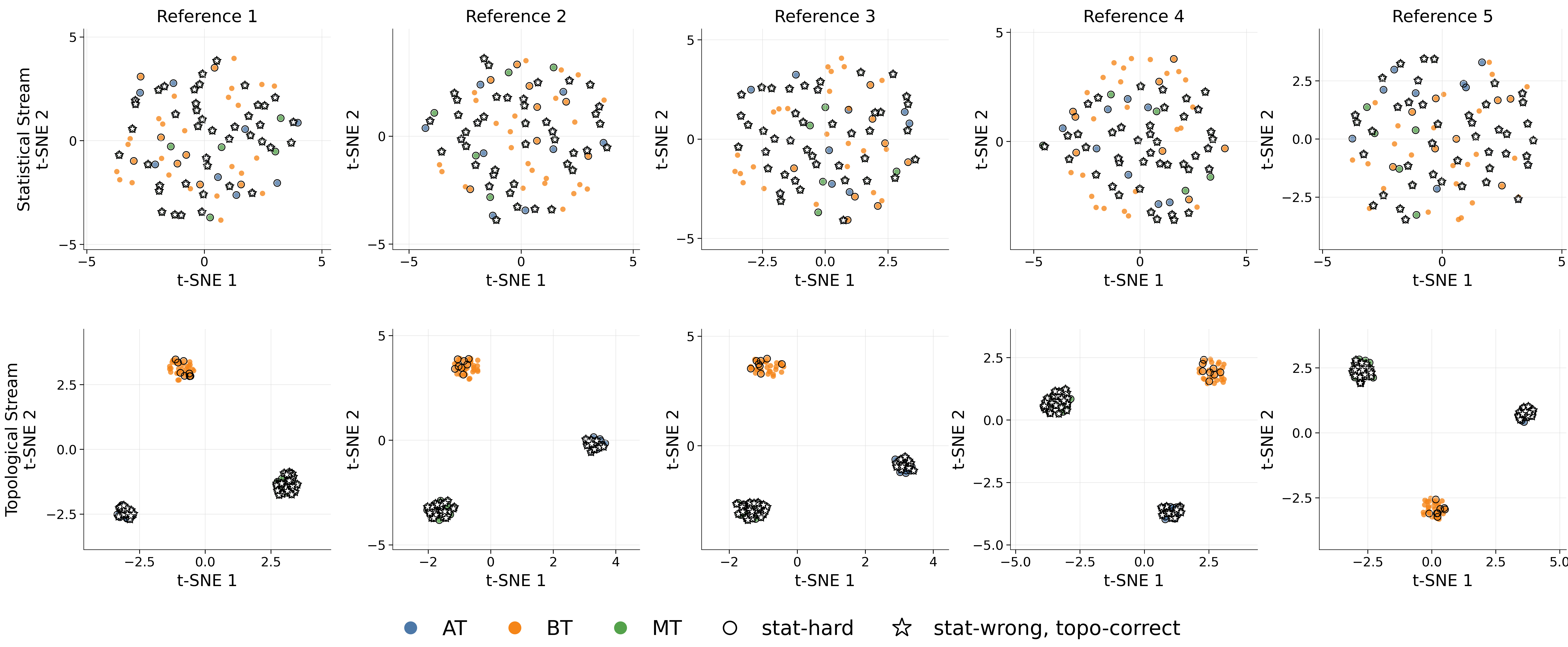}
    \caption{BRACS feature-space projections. The upper panel uses PCA and the lower uses t-SNE; each shows statistical-stream features above topological-stream features across five reference splits. Colors denote classes, hollow circles denote statistically hard samples, and stars denote cases corrected by the topological stream. Class separation is descriptive evidence, rather than a test of coordinate dependence.}
    \label{fig:feature_space_main}
\end{figure}

BRACS feature projections showed complementary class information across streams (Figure~\ref{fig:feature_space_main}). PCA silhouette scores were approximately $-0.05$ to $-0.09$ for the statistical stream and $0.87$--$0.89$ for the graph stream. The graph stream's Davies--Bouldin indices were $0.16$--$0.18$, and its t-SNE silhouettes reached $0.92$--$0.93$. Stars identify cases corrected after statistical-stream errors. These two-dimensional projections describe class separation, while perturbation tests provide the direct evidence of coordinate dependence.

The training constraint requires an additional shuffled-graph pass. Deployment constructs one intact graph, runs each stream once and adds their logits. No shuffled view, test-time ensemble or patch annotation is required. Wall-clock time, peak-memory measurements and matched-compute comparisons were unavailable, so parameter counts alone do not establish computational efficiency.
\FloatBarrier
\section{Discussion}
The central finding is that the training of a spatial pathway deserves separate evaluation from its architectural availability. The controlled task showed that changing the schedule could substantially improve learning of a spatial label within the same architecture. WSI comparisons established predictive value across several diagnostic and survival endpoints. Residual-capacity and neighborhood tests then bounded what could be attributed to spatial dependence. Together, these findings support fixing a composition reference before learning a graph correction.

The evidence also narrows the optimization interpretation. Proposition~\ref{prop:residual_gradient} gives a conditional explanation for weak supervised graph gradients when prediction error becomes small. It does not establish a universal composition-first learning trajectory. The gradient traces changed after graph training began, but the auxiliary loss also supplied gradients. Their increase cannot therefore be read as isolated recovery of spatially informative classification supervision. The matched toy and schedule comparisons provide the more direct evidence for the proposed training intervention.

Appearance capacity remains part of the residual's contribution. Coordinate-free residuals improved over the histogram anchor, showing that warm-up convergence did not exhaust the appearance multiset's predictive information. The full graph model exceeded those controls, and inference corruption reduced performance toward the anchor. Neighborhood-preserving transformations caused little change, whereas anatomical-interface rewiring produced larger losses. These observations connect prediction to supplied neighborhoods while allowing incidental correlations and unmodeled appearance information to contribute.

The approach is most relevant when spatial organization supplies information beyond composition. On a composition-dominant task, a strong appearance predictor may already be sufficient, and a graph branch could fit incidental layout. Dataset A provided a controlled negative test, but realistic composition-dominant WSI cohorts remain underrepresented. Anchor capacity and training duration remain design choices. An underfit anchor leaves appearance errors, while overfitting may weaken the useful residual-learning signal. Warm-up stability was established only for the histogram family on PANDA and BRCA; adaptive schedules and stronger appearance references require separate evaluation.

Predictive and clinical scope remain limited by the evaluated data and measurements. Only four of nine primary comparisons remained significant after multiplicity adjustment, and several alternative configurations achieved higher BRCA AUC. All WSI studies were retrospective and used frozen encoders. CAMELYON-16 supplied patch-grid localization and post-hoc anatomical evidence, rather than a segmentation system or clinical validation. Prospective cohorts, jointly tuned encoders and measured computational cost would clarify the intervention's practical limits. More realistic perturbations and composition-dominant WSI negative controls would further test when neighborhood use is beneficial.

\section{Conclusion}
ResTopoMIL addresses the gap between providing spatial inputs and learning to use them in whole-slide MIL. It fixes a composition predictor, trains an additive graph residual and separates intact from coordinate-reassigned representations. Controlled tasks support the training-schedule hypothesis, while WSI comparisons and perturbations connect predictive gains to neighborhood dependence. Coordinate-free controls retain appearance capacity as a contributing explanation. The results support explicit training interventions and behavioral spatial tests alongside architectural design.

\section*{Author Contribution Statement}
\textbf{Xiangyu Li}: Conceptualization, Methodology, Software, Investigation, Formal analysis, Visualization, Writing - original draft.
\textbf{Jincheng Li}: Conceptualization, Methodology, Supervision, Writing - review \& editing.
\textbf{Xing Xie}: Data curation, Software, Validation.
\textbf{Bin Guan}: Software, Validation.
\textbf{Shaohua Kevin Zhou}: Resources, Validation, Supervision.
\textbf{Rui Yan}: Formal analysis, Validation, Methodology, Resources.
\textbf{Ran Su}: Conceptualization, Project administration, Supervision.

\section*{Acknowledgments}
This research is supported by Open Project Program of the State Key Laboratory of MedicalProteomics(SKLP-0202406), Emerging Frontiers CultivationProgram of Tianjin University Interdisciplinary Center,SeedFoundation of Tianjin University.

\section*{Declaration}
During the preparation of this work, the authors used the Qwen model in order to refine the manuscript. After using this tool/service, the author reviewed and edited the content as needed and takes full responsibility for the content of the published article.

\bibliographystyle{elsarticle-num}
\bibliography{ref}
\end{document}